\documentclass{article}

\usepackage{arxiv}

\usepackage[utf8]{inputenc}
\usepackage[T1]{fontenc}
\usepackage{hyperref}
\usepackage{url}
\usepackage{booktabs}
\usepackage{amsfonts}
\usepackage{nicefrac}
\usepackage{microtype}
\usepackage{graphicx}
\usepackage{natbib}
\usepackage{doi}

\usepackage{subcaption}
\usepackage{array}
\usepackage{changepage}

\title{Meaning at the Planck scale? Contextualized word embeddings for doing history, philosophy, and sociology of science\\ }

\author{\href{https://orcid.org/0000-0003-0657-5254}{\includegraphics[scale=0.06]{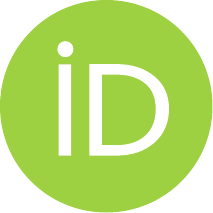}\hspace{1mm}Arno Simons} \\ History and Philosophy of Modern Science \\ Technische Universität Berlin \\ Straße des 17. Juni 135, 10623 Berlin \\ \texttt{arno.simons@gmail.com}}

\renewcommand{\headeright}{}
\renewcommand{\undertitle}{}
\renewcommand{\shorttitle}{Meaning at the Planck Scale?}

\hypersetup{
    pdftitle={Meaning at the Planck Scale? Contextualized word embeddings for doing history, philosophy, and sociology of science},
    pdfsubject={cs.CL, physics.hist-ph},
    pdfauthor={Arno Simons},
    pdfkeywords={scientific concepts, contextualized word embeddings, history of science, philosophy of science, sociology of science, word sense disambiguation and induction, lexical semantic change, discourse analysis},
}

\begin{document}
\maketitle

\begin{abstract}
This paper explores the potential of contextualized word embeddings (CWEs) as a new tool in the history, philosophy, and sociology of science (HPSS) for studying contextual and evolving meanings of scientific concepts. Using the term ``Planck'' as a test case, I evaluate five BERT-based models with varying degrees of domain-specific pretraining, including my custom model \texttt{Astro-HEP-BERT}, trained on the Astro-HEP Corpus---a dataset containing 21.84 million paragraphs from 600,000 articles in astrophysics and high-energy physics. For this analysis, I compiled two labeled datasets: (1) the Astro-HEP-Planck Corpus, consisting of 2,900 labeled occurrences of ``Planck'' sampled from 1,500 paragraphs in the Astro-HEP Corpus, and (2) a physics-related Wikipedia dataset comprising 1,186 labeled occurrences of ``Planck'' across 885 paragraphs. Results demonstrate that the domain-adapted models outperform the general-purpose ones in disambiguating the target term, predicting its known meanings, and generating high-quality sense clusters, as measured by a novel purity indicator I developed. Additionally, this approach reveals semantic shifts in the target term over three decades in the unlabeled Astro-HEP Corpus, highlighting the emergence of the Planck space mission as a dominant sense. The study underscores the importance of domain-specific pretraining for analyzing scientific language and demonstrates the cost-effectiveness of adapting pretrained models for HPSS research. By offering a scalable and transferable method for modeling the meanings of scientific concepts, CWEs open up new avenues for investigating the socio-historical dynamics of scientific discourses.
\end{abstract}


\section{Introduction}

\textbf{Scientific concepts} are dynamic entities, continually shaped and reshaped by the socio-historical contexts from which they emerge. Their meanings are influenced by the intellectual, cultural, and material conditions of their time, as well as by the disciplinary frameworks that govern their usage. This dynamic nature is evident in medical concepts such as ``syphilis'' \citep{fleck_genesis_1979} and ``atherosclerosis' \citep{mol_body_2002}, physical concepts like ``electricity'' \citep{steinle_exploratory_2016} and ``temperature'' \citep{chang_inventing_2007}, as well as more abstract methodological notions like ``probability'' \citep{hacking_emergence_1975} and ``objectivity'' \citep{daston_objectivity_2007}.

From the perspective of the \textbf{history, philosophy, and sociology of science (HPSS)}, understanding how  scientific concepts acquire and stabilize their meanings is crucial for several reasons. Changes in scientific language reflect shifts in how researchers conceptualize their subject matter, and as concepts evolve, they reshape the frameworks through which reality is interpreted. By examining these processes, HPSS highlights how scientific knowledge is not merely discovered but constructed through complex social and intellectual negotiations, emphasizing the contingent and \textbf{dynamic nature of scientific understanding} \citep{hacking_social_1999, pickering_mangle_1995, foucault_order_1994, latour_science_1987, fleck_genesis_1979, kuhn_structure_1962}. As Thomas \citet[128]{kuhn_structure_1962} famously remarked:

\begin{quote}
``the Copernicans who denied its traditional title `planet' to the sun were not only learning what `planet' meant or what the sun was. Instead, they were changing the meaning of `planet' so that it could continue to make useful distinctions in a world where all celestial bodies, not just the sun, were seen differently from the way they had been seen before''. 
\end{quote}

Transformations in the meanings of scientific concepts can also have significant societal repercussions. When concepts shift, they alter not only the internal structures of scientific knowledge but also the broader societal and cultural frameworks in which that \textbf{knowledge is embedded}. For example, historical changes in the meanings of terms like ``nature'' \citep{merchant_death_1980} or ``the normal and the pathological'' \citep{canguilhem_normal_1991} have profoundly influenced education, policy, and social hierarchy. These shifts have reinforced or challenged societal norms and power dynamics. Redefining scientific concepts creates new frameworks of understanding that affect how institutions are organized, how resources are allocated, and how humans and non-human entities are categorized and treated \citep{clarke_biomedicalization_2010, bowker_sorting_1999, hacking_social_1999}.

To study the evolving meanings of scientific concepts across different socio-historical contexts, HPSS scholars have primarily relied on qualitative methods such as \textbf{``close reading''}. While effective, these methods are often time-intensive, challenging to replicate, and limited in scalability. To complement these approaches, HPSS scholars have increasingly turned to \textbf{``distant reading''} techniques, such as co-word analysis \citep{callon_translations_1983}, latent semantic analysis (LSA) \citep{deerwester_indexing_1990}, latent Dirichlet allocation (LDA) \citep{blei_latent_2003}, and word vectors \citep{mikolov_efficient_2013}. These methods enable researchers to explore conceptual changes across large corpora, uncovering patterns in the evolution of individual terms or entire fields \citep{malaterre_epistemic_2024, lean_digital_2023, wevers_digital_2020, laubichler_computational_2019, pence_how_2018, venturini_three_2014, overton_explain_2013, boyack_clustering_2011, leydesdorff_global_2009, courtial_co-word_1989, rip_co-word_1984}. However, these approaches often struggle to capture context-dependent meanings of words and sentences.

This is where \textbf{contextualized word embeddings (CWEs)} provide a significant leap forward. CWEs rely on patterns of usage in large text corpora to encode words as points in a continuous vector space. In this space, semantic and syntactic relationships between words are represented through relative distances, with similar contexts grouped closely together. Unlike earlier word embeddings, which assign static representations to words, CWEs dynamically adjust a word's representation depending on its context. This adaptability enables researchers to analyze how meanings differ and evolve across time, disciplines, or other settings, surpassing the static associations revealed by traditional methods. 

To demonstrate the potential of CWEs for HPSS, I use \textbf{``Planck'' as a test case}, a term with multiple distinct meanings in physics---such as the physicist Max Planck, Planck units, and the Planck mission---all well-defined and widely recognized within the physics community. This diversity of known interpretations offers a clear ``ground truth'' for evaluating how effectively CWEs can capture and disambiguate these meanings, while striking a balance between simplicity and complexity. While some senses, like ``Max Planck Society'' or ``Fokker-Planck equation'', can be disambiguated through straightforward co-occurrence patterns, others demand a deeper contextual understanding. For example, domain-adapted BERT models can discern synonymous meanings in contexts such as ``a Planck function at that temperature'' and ``the spectral distribution for Planck's law'', while also interpreting nuanced cases like ``Planck's experiment yields $P_{s} \sim 10^{-9}$ for the power of scalar modes at the end of inflation'' as referring to the Planck space mission rather than Max Planck the physicist. The analysis focuses on \textbf{five BERT-based models} with varying domain specializations and training methodologies, including both general-purpose models and domain-specific models fine-tuned on scientific corpora. Among these is my own \texttt{Astro-HEP-BERT} \citep{simons_astro_2024}, a specialized model developed for conceptual analysis in astrophysics and high-energy physics (HEP), available on \href{https://huggingface.co/arnosimons/astro-hep-bert}{Hugging Face}\footnote{\url{https://huggingface.co/arnosimons/astro-hep-bert}}.

As you may have already guessed, \textbf{the title of this paper is a play on words}. In physics, the term ``Planck scale'' often refers to extremely small quantities of space and time, defined using natural units introduced by Max Planck in 1899. Here, ``meaning at the Planck scale'' serves as a metaphor for the linguistic task at hand: disambiguating the multiple senses of ``Planck''. Just as physics requires high resolution to measure reality at the smallest scales, understanding the contextual meanings of ``Planck'' requires CWEs to detect subtle semantic distinctions at the level of individual words and subwords.

The \textbf{structure of this paper} is as follows. First, I introduce CWEs as a computational tool for HPSS, highlighting their advantages over traditional methods and the benefits of BERT-based architectures. Next, I describe the datasets and labels used, providing essential context for the analysis. This is followed by a detailed comparison of the five BERT-based models used in the empirical evaluation. The core of the paper consists of four empirical sections. The first evaluates CWEs' ability to disambiguate the term ``Planck'' through supervised word-sense prediction. The second assesses the quality of sense clusters formed in an unsupervised setting, measuring alignment with labeled data. The third examines the balance between cluster separation and internal cohesion. Finally, the fourth empirical section explores the diachronic evolution of ``Planck'', tracking shifts in the prominence of different senses over time. The paper concludes by synthesizing the findings and discussing their broader implications for HPSS research. It highlights how CWEs can illuminate the socio-historical evolution of scientific concepts, identifies their limitations, and proposes future directions for integrating CWEs into the study of scientific discourses.

\section{CWEs as a new tool for HPSS}
\label{sec:cwes-as-a-new-tool-for-hpss}

CWEs represent each word in a given sequence, such as a sentence or paragraph, as a unique numerical vector in a continuous embedding space. This space encodes relationships and patterns derived from large text corpora, where words with similar meanings or contexts are positioned closer together. By dynamically adjusting a word's representation based on its specific context, \textbf{CWEs capture subtle differences in meaning}, distinguishing between various senses of the same word or phrase. For example, the word ``Planck'' in the sentence ``We used Planck's law to calculate the radiation emitted by a black body'' will have a different vector representation than in ``We used Planck's data to map the early universe'', reflecting the distinct contexts and meanings of the word in each case.

The ability to generate context-aware embeddings is a hallmark of modern transformer-based architectures, particularly \textbf{BERT (Bidirectional Encoder Representations from Transformers)} \citep{devlin_bert:_2018}. BERT's bidirectional design processes text by considering both the preceding and succeeding words in a sentence simultaneously. Through multiple layers of neural networks and attention mechanisms, it captures contextual relationships, making it especially well-suited for deep language understanding. This bidirectional approach gives BERT an edge over unidirectional models like GPT for analyzing complex linguistic contexts, though it is less effective for generative tasks.

CWEs lie at the heart of BERT's outstanding performance in natural language understanding tasks. As BERT is most often used for high-level applications like text classification, named entity recognition, or question answering, its embeddings normally remain ``under the hood''. However, \textbf{CWEs can be accessed by extracting hidden states} from BERT's intermediate or final layers. A common approach involves averaging the last four layers to create robust word embeddings \citep{periti_lexical_2024, devlin_bert:_2018}.

Though originally designed for higher-level tasks, CWEs are particularly well-suited for exploring the varying and evolving meanings of scientific concepts. Advances in \textbf{word sense disambiguation (WSD)} \citep{bevilacqua_recent_2021, loureiro_language_2020}, \textbf{word sense induction (WSI)} \citep{periti_lexical_2024, sun_method_2023}, and \textbf{lexical semantic change (LSC)} analysis \citep{periti_lexical_2024, tahmasebi_computational_2023} suggest their capacity to uncover distinct meanings that scientific terms acquire across disciplines and historical contexts \citep{zichert_tracing_2024, kleymann_conceptual_2022}. WSD identifies the intended sense of a word in context, WSI clusters word occurrences to infer senses unsupervised, and LSC analyzes shifts in word meanings over time. LSC can be considered a temporal extension of both WSD and WSI when it uses information about a word's senses to measure semantic change, for example by tracking how changes in the relative prominence of these senses reflect changes in the word's overall meaning \citep{periti_lexical_2024}.

WSD, WSI, and LSC all use BERT's ability to map semantically similar words to similar vectors, enabling the modeling of semantic relationships. These tasks largely rely on \textbf{three core operations}:

\begin{quote}
\textbf{Measuring distances between embeddings}: This operation quantifies semantic similarity or divergence between terms, typically using cosine similarity or Euclidean distance. It can be used to compare either the same term in different contexts (as in WSD and WSI) or different terms within the same context. Additionally, average pairwise similarity or distance across a set of embeddings reveals broader properties, such as the degree of polysemy, by reflecting the dispersion of meanings
\end{quote}
\begin{quote}
\textbf{Prototypical embedding creation}: By averaging the embeddings of multiple instances of a term, researchers can derive a prototypical representation of its central or typical meaning. Depending on how instances are grouped (e.g.\ by predefined sense labels for WSD, discovered clusters for WSI, or temporal periods for LSC), prototypical embeddings can reveal patterns specific to these contexts.
\end{quote}
\begin{quote}
\textbf{Clustering embeddings}: Clustering groups semantically similar terms or usages, making it valuable for WSI to uncover known or emerging senses, for WSD to generate prototypes and predict senses, and for LSC to track how senses evolve over time by grouping embeddings from different periods or contexts.
\end{quote}

Effectively using CWEs for HPSS requires understanding \textbf{domain adaptation}, which customizes BERT models for specialized contexts. BERT's pretraining relies on masked language modeling, in which certain words in a sentence are hidden and predicted based on surrounding context. This process fine-tunes the model's internal parameters through extensive training on large, diverse corpora. However, while pretrained BERT models are robust general-purpose tools, they have found to underperform in the sciences, where specialized language dominates \citep{lee_biobert_2019, beltagy_scibert_2019}. Two main strategies address this limitation: 

\begin{quote}
\textbf{Training from scratch}: A model is trained entirely on domain-specific data. This approach allows for a custom vocabulary tailored to the target domain, enhancing the model's ability to handle specialized terms. However, it requires substantial domain-specific data and significant computational resources \citep{hellert_physbert_2024, grezes_improving_2022, gu_domain-specific_2021}.
\begin{quote}
\end{quote}
\textbf{Re-using weights from pretrained models}: Pretrained models such as \texttt{BERT-base} are further trained on domain-specific corpora. This method efficiently builds on the general linguistic knowledge captured during pretraining, adapting the model to domain-specific terminology and contexts. Although this approach is resource-efficient, it typically reuses the original vocabulary, which may limit the model's ability to fully represent highly specialized terms. Subword tokenization helps mitigate this issue but cannot entirely address it \citep{devlin_bert:_2018}.
\end{quote}

The choice between these approaches depends on the \textbf{availability of data and computational resources}. Additional training of a pretrained model is often sufficient for many applications---as demonstrated in this paper by the comparative performance of \texttt{Astro-HEP-BERT} against models like \texttt{PhysBERT} and \texttt{astroBERT}, which were trained from scratch---while training from scratch may be necessary for highly specialized tasks that demand more precise domain knowledge \citep{hellert_physbert_2024, grezes_improving_2022, grezes_building_2021}.

\section{Datasets and labels}
\label{sec:datasets-and-labels}

For this study, I use two custom-compiled datasets. 

\begin{quote}
The \href{https://huggingface.co/datasets/arnosimons/astro-hep-corpus}{\textbf{Astro-HEP Corpus}}\footnote{\url{https://huggingface.co/datasets/arnosimons/astro-hep-corpus}} consists of 21.84 million paragraphs extracted from 600,000 scientific articles (approximately 4.2 billion tokens) published on arXiv between 1986 and 2022, spanning the domains of astrophysics and high-energy physics (HEP). The paragraphs were extracted in plain text format from the original LaTeX files using \href{https://pandoc.org/}{Pandoc}\footnote{\url{https://pandoc.org/}}. To ensure data quality, I replaced literature references with ``[CIT]'' and multiline formulas with ``FORMULA''. Additionally, paragraphs that were either unusually short or exhibited an abnormal frequency of whitespace characters were filtered out based on character frequency analysis.
\end{quote}
\begin{quote}
The \href{https://huggingface.co/datasets/arnosimons/wikipedia-physics-corpus}{\textbf{Wikipedia-Physics Corpus}}\footnote{\url{https://huggingface.co/datasets/arnosimons/wikipedia-physics-corpus}} contains 102,409 paragraphs extracted from 6642 key physics-related Wikipedia articles. These articles were selected using the \href{https://petscan.wmcloud.org/}{PetScan}\footnote{\url{https://petscan.wmcloud.org/}} tool, which generated a list of all pages categorized under ``physics'' or its immediate subcategories. Markup was removed and minimal cleaning applied to produce plain text paragraphs, and while references were removed, all formulas were retained.
\end{quote}

To establish a \textbf{ground truth} for the WSD and LSC analyses of ``Planck'' (Sections \ref{sec:word-sense-prediction}-\ref{sec:lexical-semantic-change}), I annotated occurrences of the term in both corpora using \textbf{21 predefined labels}, determined by theoretical insights, expert knowledge, and a qualitative review of the contexts. In the Astro-HEP Corpus, I manually labeled 2,900 occurrences from a random sample of 1,500 paragraphs, and this subset is referred to as the Astro-HEP-Planck Corpus. In the Wikipedia-Physics Corpus, I labeled all 1,186 occurrences across 885 paragraphs. I identified occurrences using a case-insensitive regular expression to capture variations like ``PLANCK'' and ``Planck(2015)'' while excluding irrelevant forms like ``planckian''. 

\begin{figure}[htbp!]
  \centering
  \begin{subfigure}[t]{.33\linewidth}
    \centering\includegraphics[width=1\linewidth]{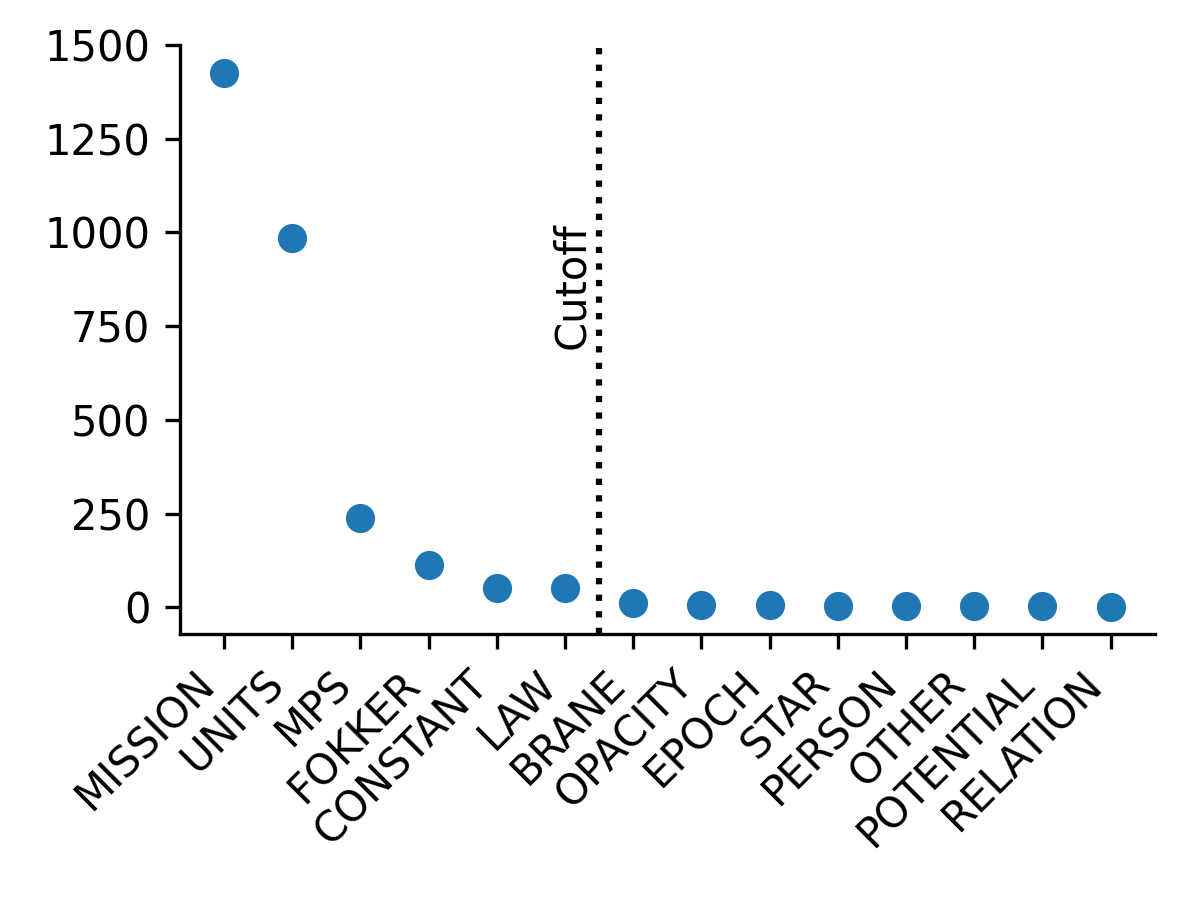}
    \caption{Astro-HEP-Planck Corpus}
  \end{subfigure}
  \begin{subfigure}[t]{.5\linewidth}
    \centering\includegraphics[width=1\linewidth]{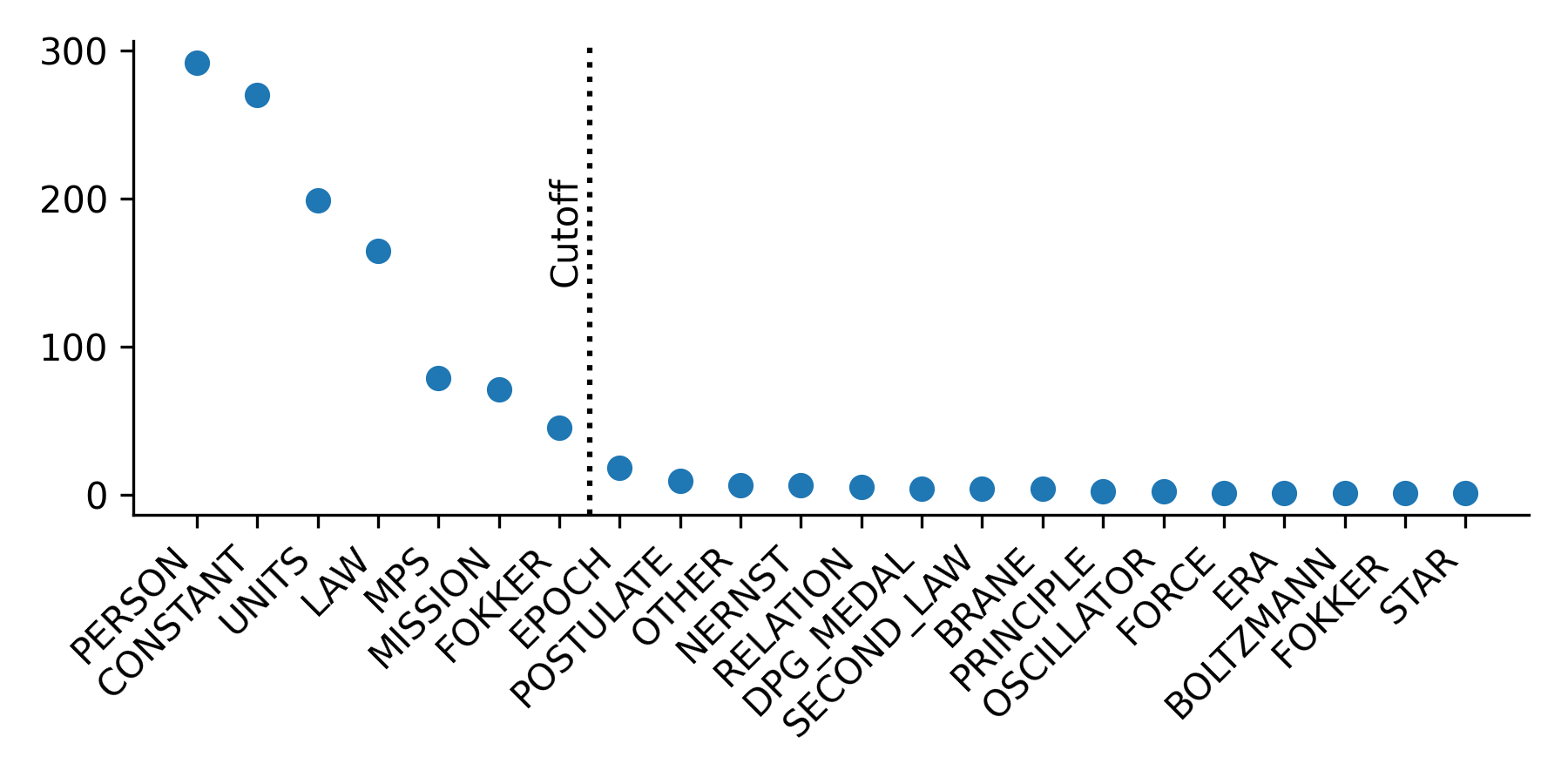}
    \caption{Wikipedia-Physics Corpus}
  \end{subfigure}
  \caption{Distribution of labels and their cutoff thresholds for the word sense disambiguation and induction tasks in the Astro-HEP-Planck and Wikipedia-Physics corpora. The x-axis shows the labels ranked by frequency, while the y-axis represents the number of occurrences. Cutoffs indicate the number of labels included in each subset for subsequent analyses.}
  \label{fig:label-distribution}
\end{figure}

For further analysis, I focused on occurrences corresponding to the \textbf{seven most common labels} (see Table \ref{table:label-description} and Figure \ref{fig:label-distribution}). For each corpus, I created \textbf{six subsets} containing the two, three, four, five, six, and seven most common labels within the corpus, respectively. For each subset, I extracted all CWEs by averaging the model's final four hidden layers \citep{periti_lexical_2024, devlin_bert:_2018}.

\begin{table}[htbp!]
\centering
\begin{tabular}{>{\raggedright\arraybackslash}p{1.4cm}%
   >{\raggedleft\arraybackslash}p{0.7cm}%
   >{\raggedleft\arraybackslash}p{0.7cm}%
   >{\raggedright\arraybackslash}p{12cm}%
  }
 \toprule
 Label & AHC & WPC & Description \\ 
 \midrule
 \textit{CONSTANT} & 52 & 270 & Planck's constant $h$: A fundamental physical constant of foundational importance in quantum mechanics. A photon's energy is equal to its frequency multiplied by the Planck constant, and the wavelength of a matter wave equals the Planck constant divided by the associated particle momentum.\\
 \textit{FOKKER} & 112 & 45 & Fokker–Planck equation: A partial differential equation that describes the time evolution of the probability density function of the velocity of a particle under the influence of drag forces and random forces, as in Brownian motion.\\
 \textit{LAW} & 52 & 165 & Planck's law: A formula that describes the spectral density of electromagnetic radiation emitted by a black body in thermal equilibrium at a given temperature, when there is no net flow of matter or energy between the body and its environment.\\
 \textit{MISSION} & 1426 & 71 & Planck mission: A space observatory operated by the European Space Agency from 2009 to 2013. The project that mapped the anisotropies of the cosmic microwave background (CMB) at microwave and infrared frequencies, with high sensitivity and angular resolution.\\
 \textit{MPS} & 239 & 79 & The Max Planck Society for the Advancement of Science: A formally independent non-governmental and non-profit association of German research institutes.\\
 \textit{PERSON} & 4 & 292 & Max Planck: The famous German theoretical physicist whose discovery of energy quanta won him the Nobel Prize in Physics in 1918.\\
 \textit{UNITS} & 986 & 199 & Planck units: A system of units of measurement defined exclusively in terms of four universal physical constants: $c$, $G$, $\hbar$, and $kB$. Expressing one of these physical constants in terms of Planck units yields a numerical value of 1.\\
  \bottomrule
  \hfill
\end{tabular}
\caption{Description of the seven most common labels (in alphabetical order) used for word sense disambiguation of the term ``Planck''. Columns AHC and WPC depict the number of labeled occurrences in the Astro-HEP-Planck Corpus and the Wikipedia-Physics Corpus, respectively.}
\label{table:label-description}
\end{table} 

\section{Models}
\label{sec:models}

I compare the performance of five BERT-based models: The first model, \texttt{BERT-base} or simply \textbf{BERT} \citep{devlin_bert:_2018}, is a general-purpose model trained from scratch over 40 epochs on a dataset of Wikipedia articles and free books, totaling 3.3 billion tokens. It uses a 30,000-token uncased vocabulary and serves as the baseline model for this analysis. The second model, \texttt{SciBERT-scivocab-uncased} or simply \textbf{SciBERT} \citep{beltagy_scibert_2019}, was developed specifically for scientific text and trained from scratch on 1.14 million biomedical and computer science papers, with its uncased vocabulary learned directly from the corpus. The third model, \textbf{astroBERT} \citep{grezes_improving_2022}, was trained on 400,000 astrophysics publications from the SAO/NASA Astrophysics Data System (ADS), specifically designed to identify, disambiguate, and tag entities in astrophysical literature. Its vocabulary is case-sensitive and was learned from the corpus. The fourth model, \texttt{PhysBERT-uncased} or simply \textbf{PhysBERT} \citep{hellert_physbert_2024}, was trained from scratch over 10 epochs on 1.2 million arXiv physics papers, with its uncased vocabulary learned directly from the corpus. The model was also fine-tuned to produce domain-adapted sentence embeddings.

The fifth model, \textbf{Astro-HEP-BERT}, is my own \citep{simons_astro_2024}. Instead of training it from scratch, as was done for the other four models, I reused \texttt{BERT}'s learned linguistic patterns and vocabulary and retrained the model for three additional epochs using on my Astro-HEP Corpus. To enhance learning, I trained the model solely on entire paragraphs rather than packing in as many sentences as possible, as often suggested in BERT tutorials. This ``full-paragraphs format'' preserves sentences within their original context, which is especially meaningful in academic writing where paragraphs focus on one idea. Previous research shows that using multiple sentences from the same document improves results \citep{liu_roberta_2019}, so I expected this approach to help the model capture more semantic information. 

\section{Word sense prediction using sense prototypes}
\label{sec:word-sense-prediction}

As a first WSD test I use a simple supervised \textbf{1-nearest neighbor (1NN)} approach as described in \citep{loureiro_language_2020}. This test compares the models' ability to predict the correct sense of ``Planck'' occurrences by computing the cosine similarity between their CWEs and \textbf{precomputed sense prototypes}. The closest prototype is then used to assign the appropriate sense to the target occurrence. Failure to classify an occurrence correctly suggests that the CWEs for the senses are not well-separated. For each corpus, the sense prototypes are generated by averaging the CWEs of all labeled ``Planck'' occurrences corresponding to each sense.

\begin{figure}[htbp!]
  \centering
  \begin{subfigure}[t]{.35\linewidth}
    \centering\includegraphics[width=1\linewidth]{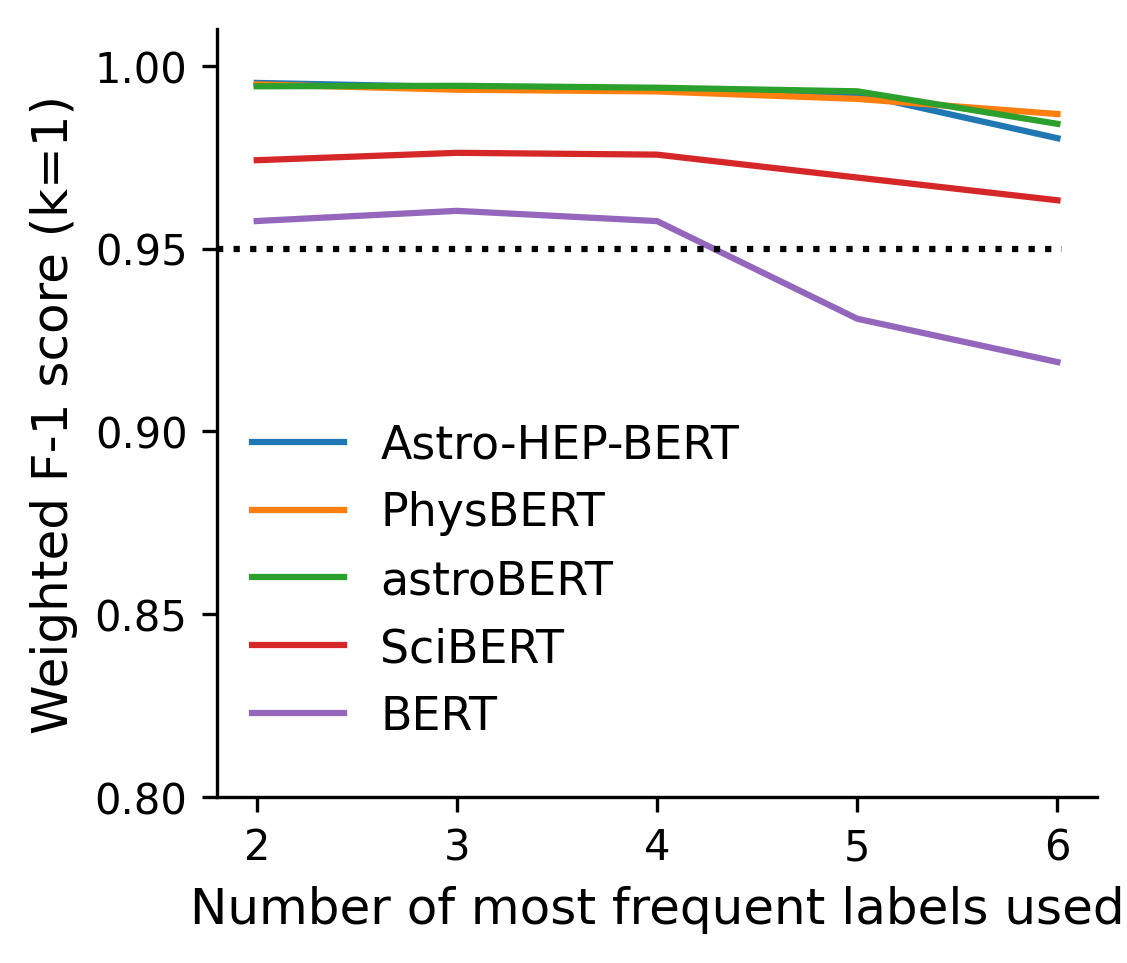}
    \caption{Astro-HEP-Planck Corpus}
  \end{subfigure}
  \begin{subfigure}[t]{.335\linewidth}
    \centering\includegraphics[width=1\linewidth]{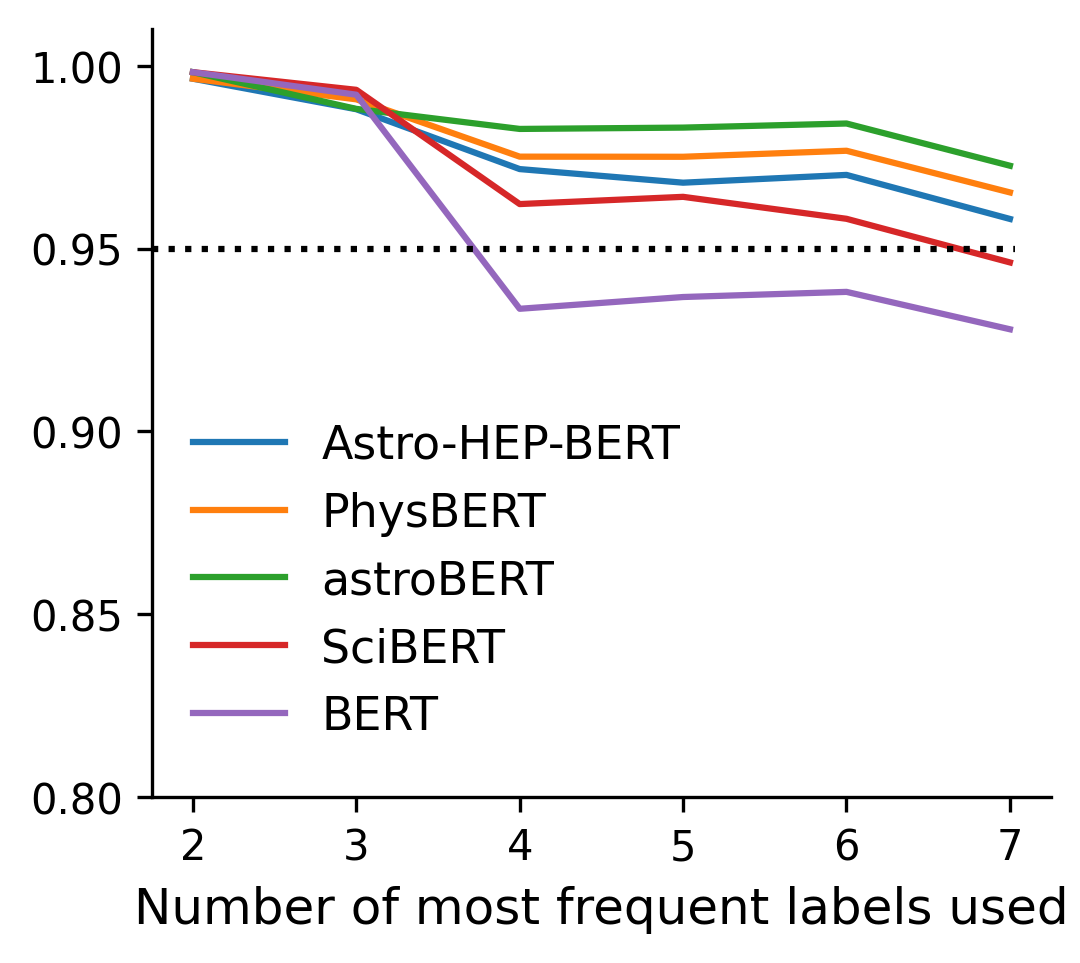}
    \caption{Wikipedia-Physics Corpus}
  \end{subfigure}
  \caption{Comparison of model performance in disambiguating the word ``Planck'' using $1$-nearest neighbor (1NN) classifiers across subsets of labels from the Astro-HEP-Planck and Wikipedia-Physics corpora. The x-axis represents the number of labels in each subset, while the y-axis shows the weighted F-1 scores. Each model was evaluated on classifiers trained on CWEs extracted for two to six labels in the Astro-HEP-Planck Corpus and two to seven labels in the Wikipedia-Physics Corpus.}
  \label{fig:knn-classification}
\end{figure}

Figure \ref{fig:knn-classification} compares the models' performance in predicting different senses of ``Planck'' using word sense prototypes across the two corpora. Each model was evaluated using six separate $1$NN classifiers, with each classifier trained on a subset containing between two and seven of the most common labels in the corpus. Figures \ref{fig:knn-classification}a and \ref{fig:knn-classification}b present the weighted F-1 scores for the Astro-HEP-Planck and Wikipedia-Physics corpora, respectively. The x-axis represents the number of labels in the subset, while the y-axis shows the F-1 score for each model. This setup enables a direct comparison of how effectively the models distinguish between the senses of ``Planck'' as the classification task becomes more complex with the inclusion of additional senses (labels).

In both figures, we observe that the classifiers generally achieve higher F-1 scores when they have to predict fewer labels, with performance declining as the number of labels---and thus ambiguity---increases. This in itself should not be surprising because fewer labels make the classification task simpler, reducing ambiguity, whereas more labels increase complexity and the likelihood of misclassification. A minor exception occurs in the Wikipedia-Physics Corpus, where all models show a slight performance rebound after an initial decline when predicting the four most common labels. The initial decline in the four-label case is primarily due to misclassifying CWEs labeled as \textit{LAW}---the newly introduced fourth most frequent label---as either \textit{CONSTANT} or \textit{PERSON}, and vice versa. As the labels introduced in the five-, six-, and seven-label settings are less ambiguous, the overall F-1 scores improve slightly.

The results also highlight differences in how well the models handle the challenge of dealing with increasing numbers of labels and ambiguity: some maintain relatively high performance even as the number of labels grows, while others experience a steeper decline. Overall, \texttt{PhysBERT}, \texttt{astroBERT}, and \texttt{Astro-HEP-BERT} perform exceptionally well on the Astro-HEP-Planck Corpus, consistently achieving F-1 scores above 0.98, with most scores exceeding 0.99. \texttt{SciBERT} follows, with F-1 scores generally around 0.97. The weakest performer in the Astro-HEP-Planck Corpus is \texttt{BERT}, starting with an F-1 score near 0.96 and dropping to 0.92 in the six-label case. In the Wikipedia-Physics Corpus, \texttt{astroBERT} outperforms the other models, maintaining F-1 scores above 0.97. \texttt{PhysBERT} and \texttt{Astro-HEP-BERT} follow closely, with scores above 0.96 and 0.95, respectively. Although \texttt{BERT} starts strong with F-1 scores near 0.99 in the two- and three-label cases, its performance declines sharply, dropping to around 0.93 in the more complex label settings.

While \texttt{BERT} shows the weakest performance across both corpora, its F-1 score never falls below 0.92, reflecting the strong baseline performance of BERT models in word sense disambiguation \citep{wiedemann_does_2019}. In contrast, the stronger performance of \texttt{astroBERT}, \texttt{PhysBERT}, and \texttt{Astro-HEP-BERT} in disambiguating the term ``Planck'' may suggest the advantages of domain-specific pretraining.

The results of the word sense prediction task showed that domain-specific models outperform general-purpose ones in distinguishing between distinct senses of ``Planck''. However, sense disambiguation often requires unsupervised approaches, especially when sense labels are unknown. The next section evaluates the models' capacity to cluster senses effectively without supervision.

\section{Cluster purity}
\label{sec:cluster-purity}

While the previous section evaluated the models' ability to classify ``Planck'' occurrences into predefined senses, a classical WSD task, this section evaluates the models' ability to perform WSI by forming meaningful word sense clusters in an unsupervised manner employing the \textbf{K-means clustering} \citep[cf.][]{periti_lexical_2024, sun_method_2023}. This algorithm partitions data into a predefined number of $k$ clusters by initializing centroids (usually randomly) and iteratively refining them to minimize variance within each cluster, based on the average distance between data points and their nearest centroid. For each corpus, model, and subset of labeled CWEs, $k$ is set to the number of labels $l$ in the subset. For each value of $k$, the best clustering solution out of 100 random seed initializations is selected based on inertia---the sum of squared distances between CWEs and their nearest cluster center.

To evaluate how effectively a clustering solution $s$ disambiguated the predefined senses of ``Planck'', I propose a new \textbf{purity indicator}, using a coefficient of variation (CV) approach, where I compare the weighted average CV for the clusters against a theoretical maximum that can occur if clusters are perfectly pure. The CV is a standardized measure of dispersion of frequency distribution, defined as the ratio of the standard deviation $\sigma$ to the mean $\mu$. Building on this definition, my purity indicator is computed as:

$$\mbox{purity}_s=\frac{\mbox{waCV}_s}{\mbox{TM}_l},$$ 

where $\mbox{waCV}_s$ is the weighted average CV for the clusters in solution $s$, and $\mbox{TM}_l$ the theoretical maximum value that any empirical $\mbox{waCV}_s$ can attain, given $l$ available labels. The $\mbox{waCV}_s$ is computed as:

$$\mbox{waCV}_s = \sum \limits_{i=1}^{k}w_{i}'\mbox{CV}_{i}$$

with $k$ as the number of clusters, $w_i'$ the normalized size of cluster $i$, and $\mbox{CV}_i$ the CV of the label distribution in cluster $i$, defined as:

$$\mbox{CV}_i = \frac{\sigma_i}{\mu_i},$$

where $\sigma_i$ is the standard deviation and $\mu_i$ the mean of the label distribution in cluster $i$. The theoretical maximum $\mbox{TM}_l$ for $\mbox{waCV}_s$, given $l$ labels, occurs when all clusters are ``pure'', meaning they contain only one type of label. Thus, we can simplify to a single-cluster-single-label scenario:

$$\mbox{TM}_l=\frac{\sigma_f((1,0^{l-1}))}{\mu_f((1,0^{l-1})))}$$

A high purity score indicates that clusters are mostly composed of instances carrying the same label, whereas a lower score suggests mixing of labels, indicating less effective disambiguation.

\begin{figure}[htbp!]
  \centering
  \begin{subfigure}[t]{.463\linewidth}
    \centering\includegraphics[width=1\linewidth]{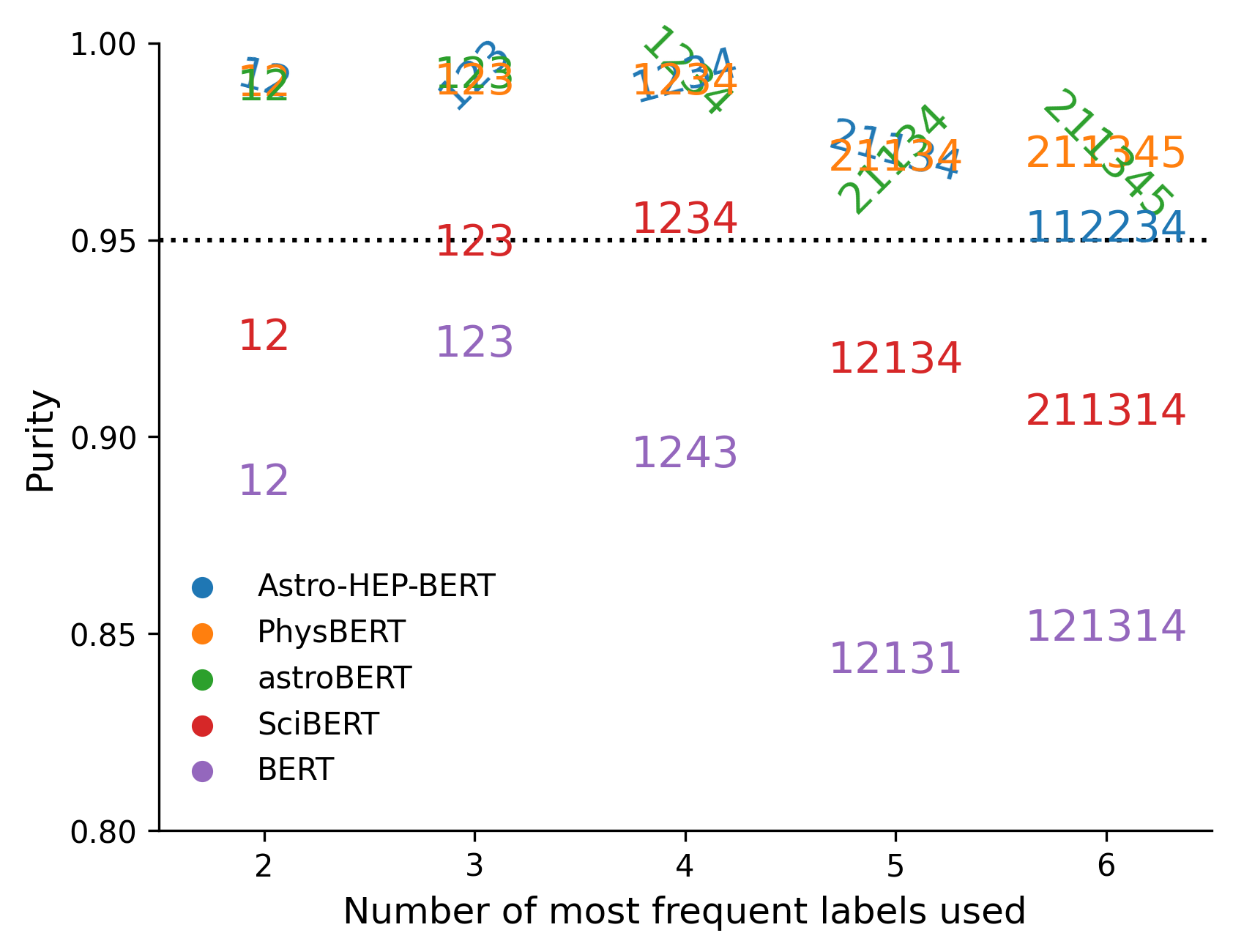}
    \caption{Astro-HEP-Planck Corpus}
  \end{subfigure}
  \begin{subfigure}[t]{.532\linewidth}
    \centering\includegraphics[width=1\linewidth]{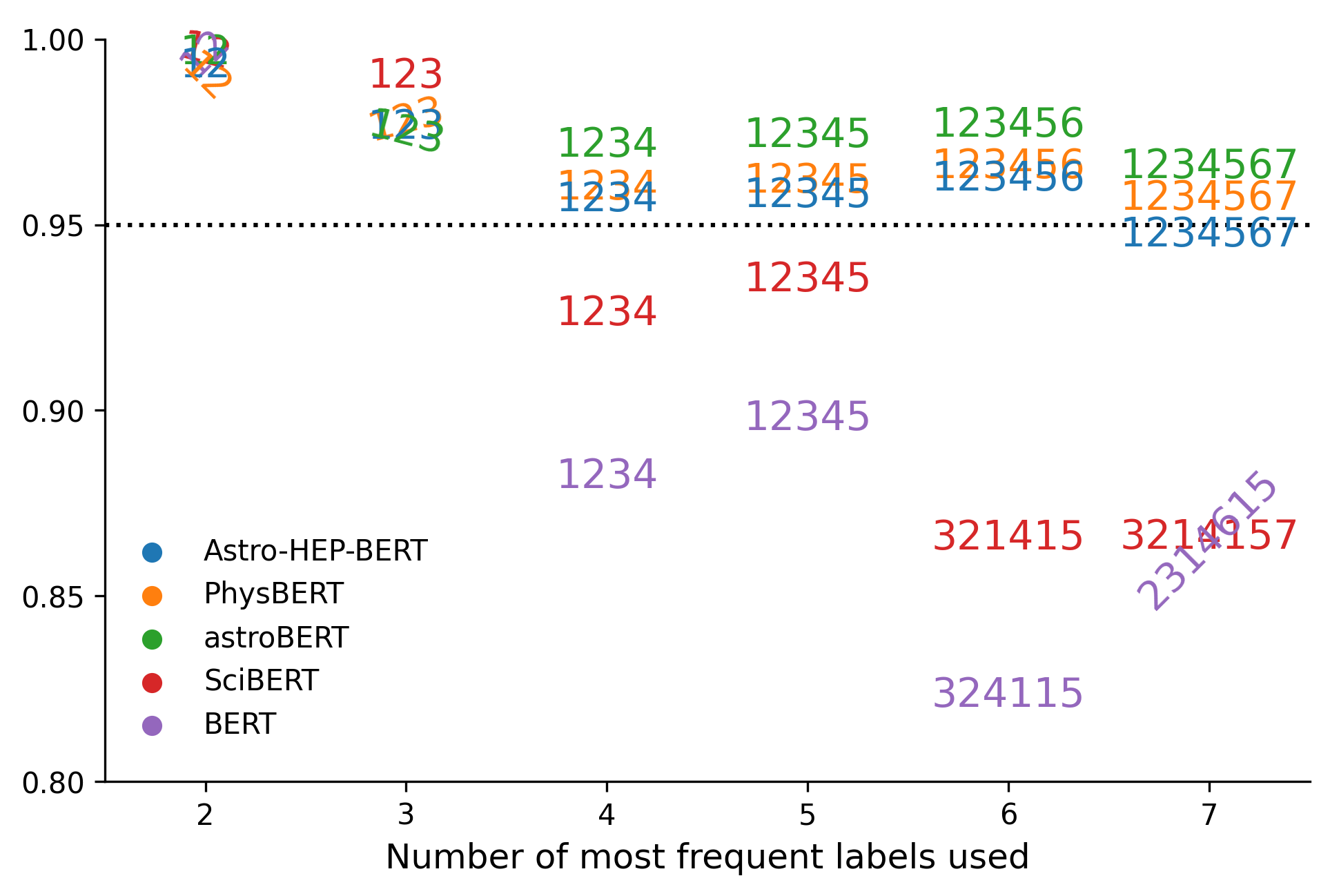}
    \caption{Wikipedia-Physics Corpus}
  \end{subfigure}
  \caption{Purity scores of clustering solutions across models and subsets of CWEs for the Astro-HEP-Planck and Wikipedia-Physics corpora. The x-axis represents the number of predefined word sense labels in each subset, and the y-axis shows the purity scores for each model. Clustering solutions are annotated with permutations of dominant labels, reflecting the frequency distribution of word senses in the datasets. Detailed label mappings can be inferred from figures \ref{fig:label-distribution}a and \ref{fig:label-distribution}b for the Astro-HEP-Planck and Wikipedia-Physics corpora, respectively.}
  \label{fig:cluster-comparison}
\end{figure}

Figures \ref{fig:cluster-comparison}a and \ref{fig:cluster-comparison}b display the purity scores of cluster solutions across models and subsets for the Astro-HEP-Planck and Wikipedia-Physics corpora. The x-axis represents the subset label count, while the y-axis shows the purity score. Each cluster solution is annotated with a permutation of dominant labels, reflecting their frequency in the corpus. For instance, ``1234'' represents clusters dominated by the top four labels in descending order of frequency. By contrast, ``1243'' (seen in \texttt{BERT}'s four-label solution for the Astro-HEP-Planck Corpus) indicates that the third-largest cluster is dominated by the fourth label rather than the third, suggesting an artificial size increase due to ``pollution'' from other labels.

Results indicate distinct model performance patterns. In the Astro-HEP-Planck Corpus, \texttt{Astro-HEP-BERT}, \texttt{PhysBERT}, and \texttt{astroBERT} achieve the highest clustering quality, consistently above 0.95, and even exceed 0.99 with fewer labels. By contrast, \texttt{SciBERT} and \texttt{BERT} peak at 0.96 and 0.92, respectively, with performance declining faster as label complexity increases. Up to four labels, most models maintain the corpus's frequency order, with \texttt{BERT} as an exception. In the five- and six-label cases, however, all models deviate from frequency order, indicating challenges as complexity rises. Here, \texttt{Astro-HEP-BERT}, \texttt{PhysBERT}, and \texttt{astroBERT} tend to form multiple clusters dominated by labels like \textit{MISSION} or \textit{UNITS}, indicating potential context-specific subgroups within dominant senses. \texttt{SciBERT} and \texttt{BERT} also show overrepresentation, producing up to three \textit{MISSION}-dominant clusters in the five- and six-label cases.

These trends may stem from domain-specific pretraining in \texttt{Astro-HEP-BERT}, \texttt{PhysBERT}, and \texttt{astroBERT}, enhancing their sensitivity to physics terms like ``Planck''. Conversely, \texttt{SciBERT} and \texttt{BERT} lack this focus and may thus merge distinct senses. Additionally, the K-Means algorithm, designed to minimize variance around centroids, may contribute to overrepresenting dominant senses by favoring larger clusters around common labels, often at the expense of rarer ones. Thus, beyond model-specific factors, the clustering method itself may introduce a bias toward frequent senses, influencing observed patterns

In the Wikipedia-Physics Corpus, \texttt{astroBERT} consistently outperforms, maintaining purity scores above 0.97, with \texttt{PhysBERT} and \texttt{Astro-HEP-BERT} closely following. \texttt{SciBERT} achieves the highest purity (0.99) in the three-label scenario, but its performance drops in six- and seven-label cases, while \texttt{BERT} shows inconsistent results. Models perform well in two- to five-label cases, aligning closely with corpus frequency order. Beyond five labels, deviations appear, particularly for \texttt{SciBERT} and \texttt{BERT}, which form multiple clusters dominated by the \textit{PERSON} label in the six- and seven-label scenarios, highlighting challenges with less frequent senses.

Overall, \texttt{astroBERT}, \texttt{PhysBERT}, and \texttt{Astro-HEP-BERT} demonstrate superior clustering quality, while \texttt{BERT} and \texttt{SciBERT} fall behind. Although the top three models maintain high purity scores across both corpora, they more accurately reproduce the label distribution in the Wikipedia-Physics Corpus, likely due to its more balanced label frequencies (see Figure \ref{fig:label-distribution}), more conventional language, and clearer semantic boundaries. In contrast, the specialized Astro-HEP-Planck Corpus, with its technical language and niche contexts, seems to present challenges for all models and for K-Means in accurately distinguishing less frequent senses.

\section{Cluster separation and cohesion}
\label{sec:separation-and-cohesion}

In addition to purity, I evaluate the balance between internal cluster cohesion and inter-cluster separation to better assess the quality of clustering solutions across different models, particularly in cases where purity scores were similar. To measure inter-cluster separation, I calculate the \textbf{average pairwise similarity (APS)} between CWEs from different clusters:

$$\mbox{APS}(E_1, E_2) = \frac{1}{|E_1||E_2|} \sum \limits_{e_{1, i} \in E_1, e_{2, j} \in E_2} \mbox{CS}(e_{1, i}, e_{2, j}),$$

where $E_1$ and $E_2$ are the sets of CWEs in two different clusters, and $\mbox{CS}$ is the cosine similarity function quantifying similarity between pairs of CWEs from different clusters.

For internal cluster cohesion, I compute the \textbf{average inner similarity (AIS)} within each cluster:

$$\mbox{AIS}(E, E) = \frac{2}{|E|(|E|-1)} \sum \limits_{e_i, e_j \in E, e_i \neq e_j} \mbox{CS}(e_i, e_j),$$

where $E$ represents the set of CWEs within a single cluster. 

These metrics provide insight into how well-separated the clusters are and how cohesive each cluster is internally. Low APS scores suggest clear distinctions between clusters, while high AIS scores indicate strong internal coherence within clusters. 

\begin{figure}
\centering
    \begin{subfigure}[t]{.49\linewidth}
        \centering\includegraphics[width=1\linewidth]{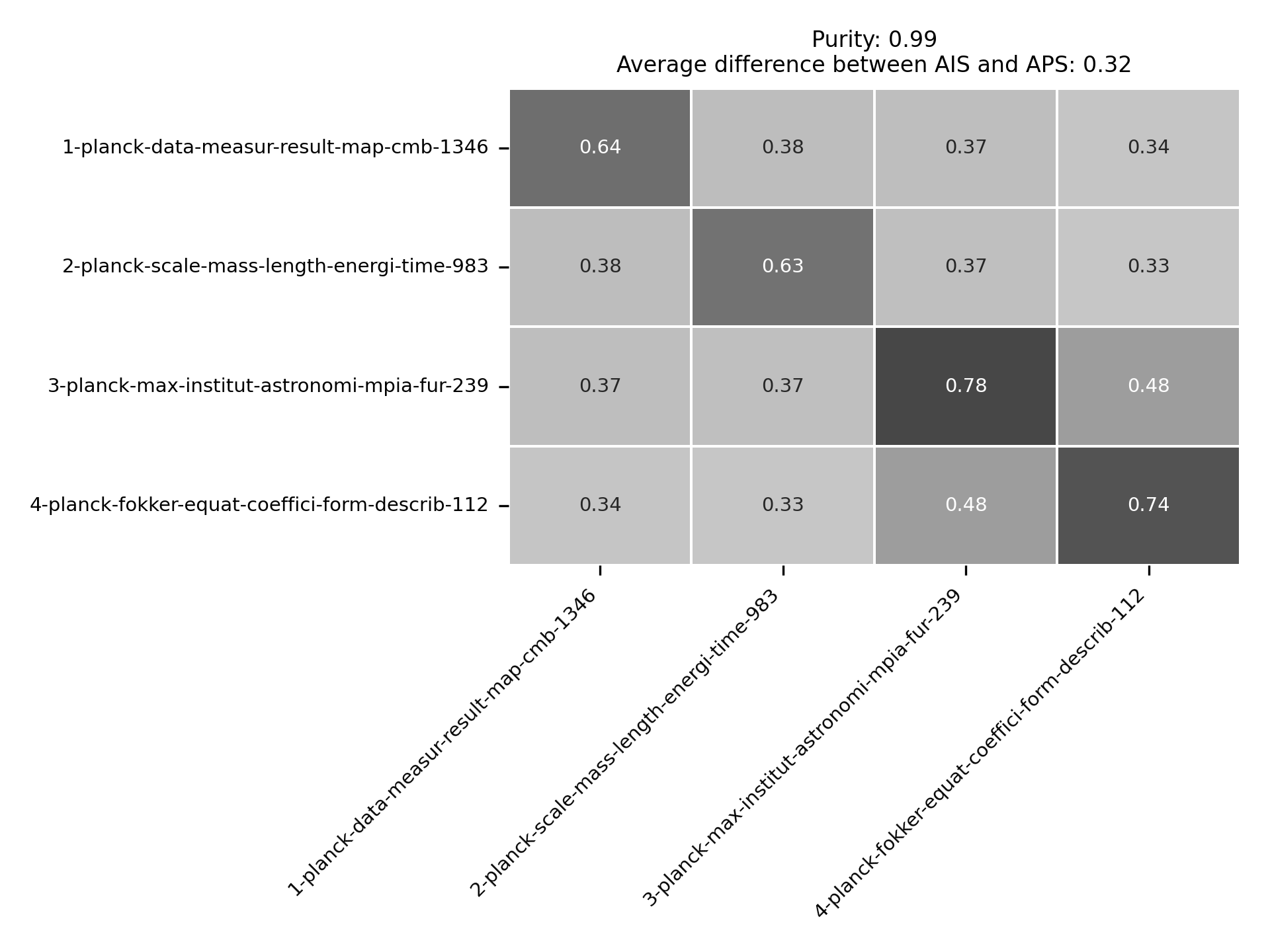}
        \caption{Astro-HEP-BERT}
    \end{subfigure}
    \begin{subfigure}[t]{.49\linewidth}
        \centering\includegraphics[width=1\linewidth]{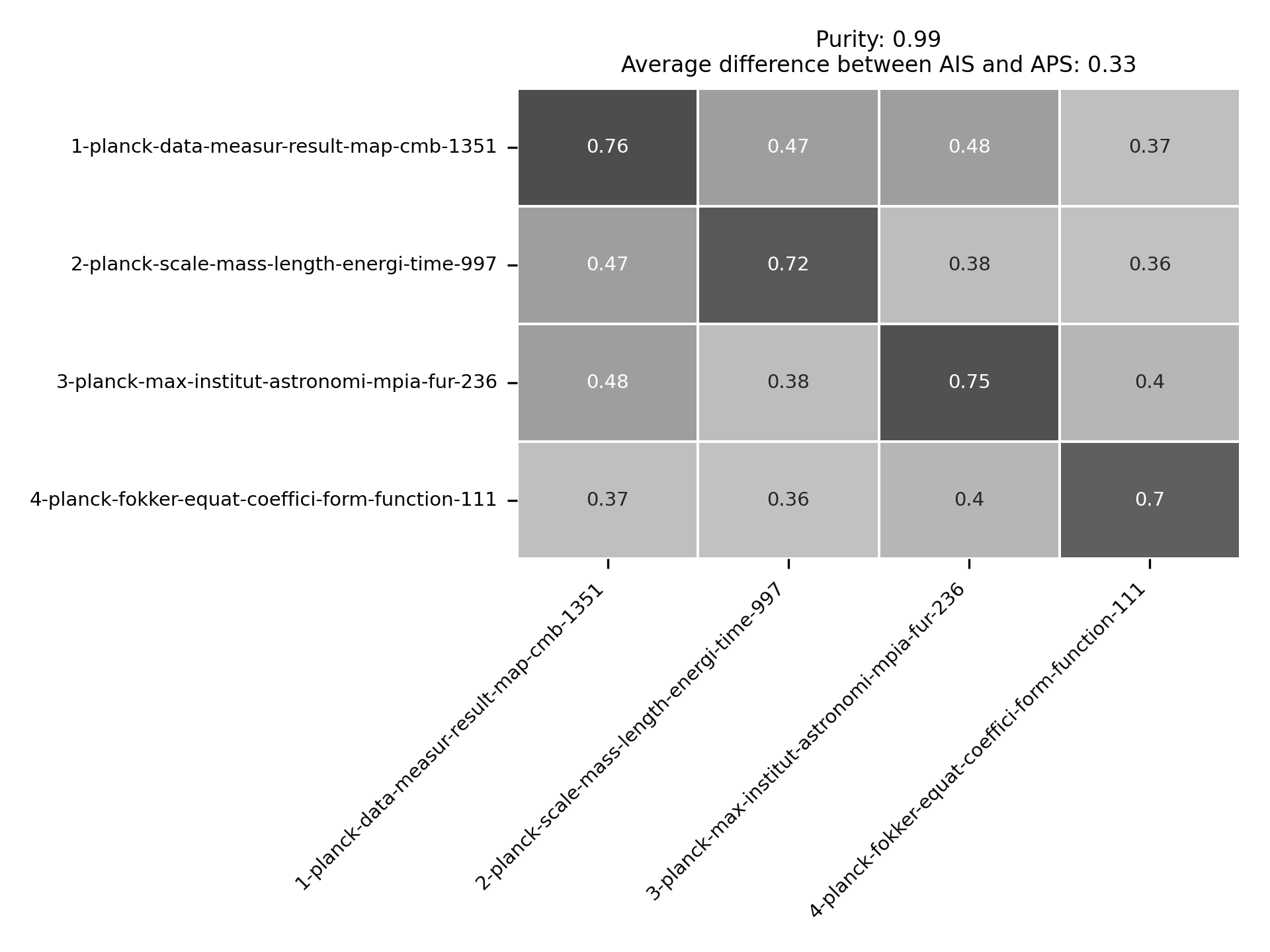}
        \caption{PhysBERT}
    \end{subfigure}
    \par\bigskip
    \begin{subfigure}[t]{.49\linewidth}
        \centering\includegraphics[width=1\linewidth]{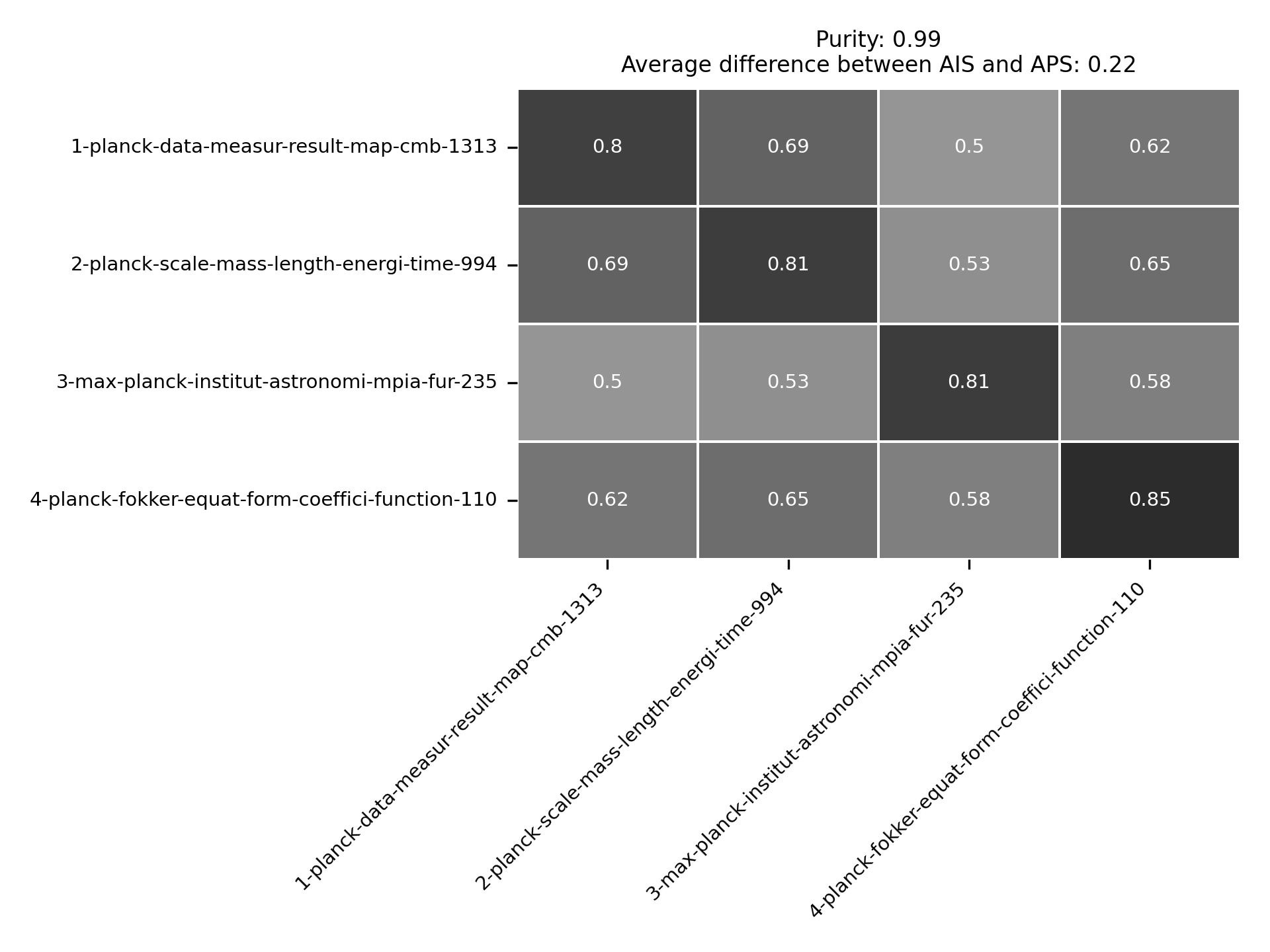}
        \caption{astroBERT}
    \end{subfigure}
    \begin{subfigure}[t]{.49\linewidth}
        \centering\includegraphics[width=1\linewidth]{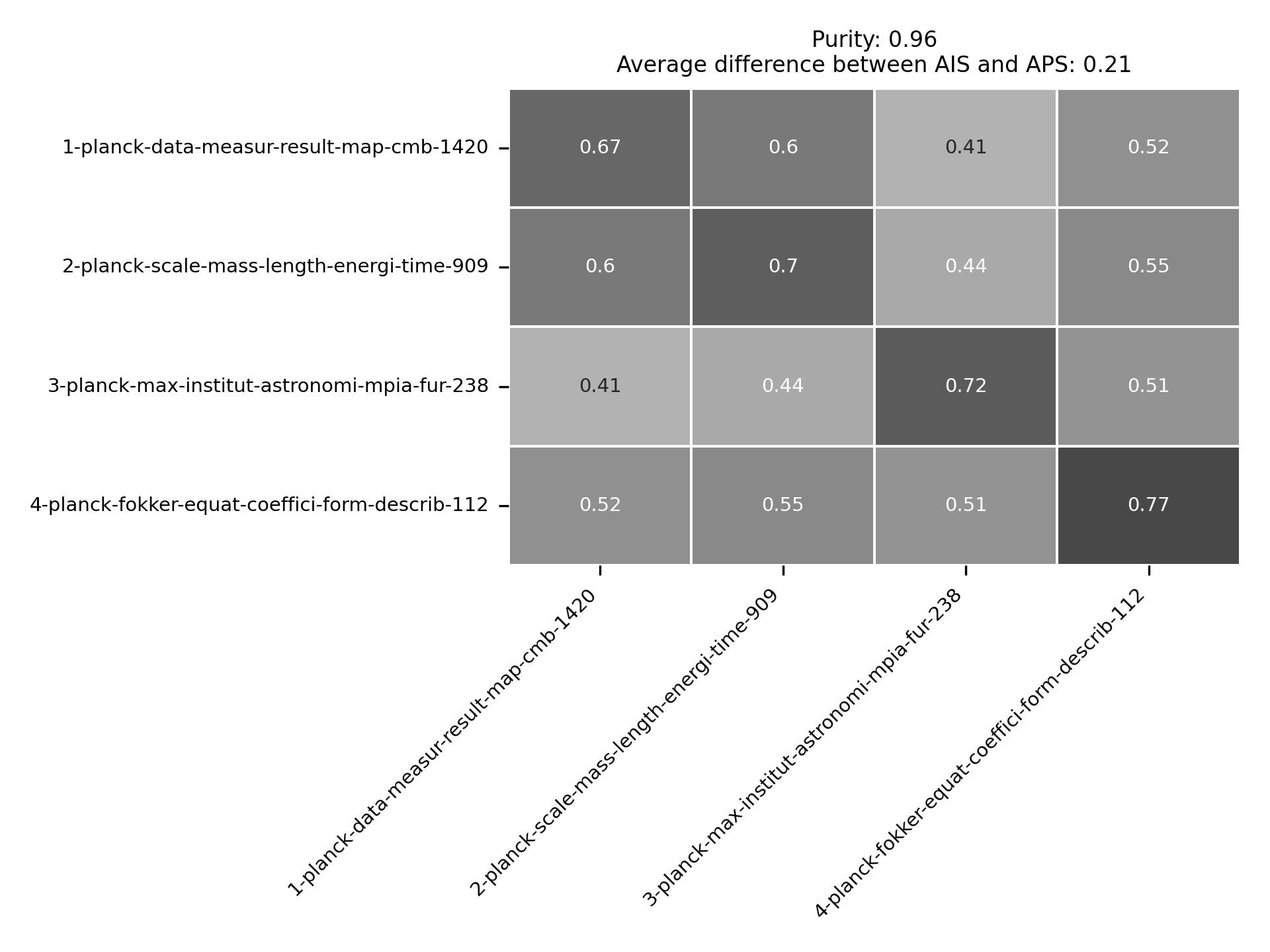}
        \caption{SciBERT}
    \end{subfigure}
    \par\bigskip
    \begin{subfigure}[t]{.49\linewidth}
        \centering\includegraphics[width=1\linewidth]{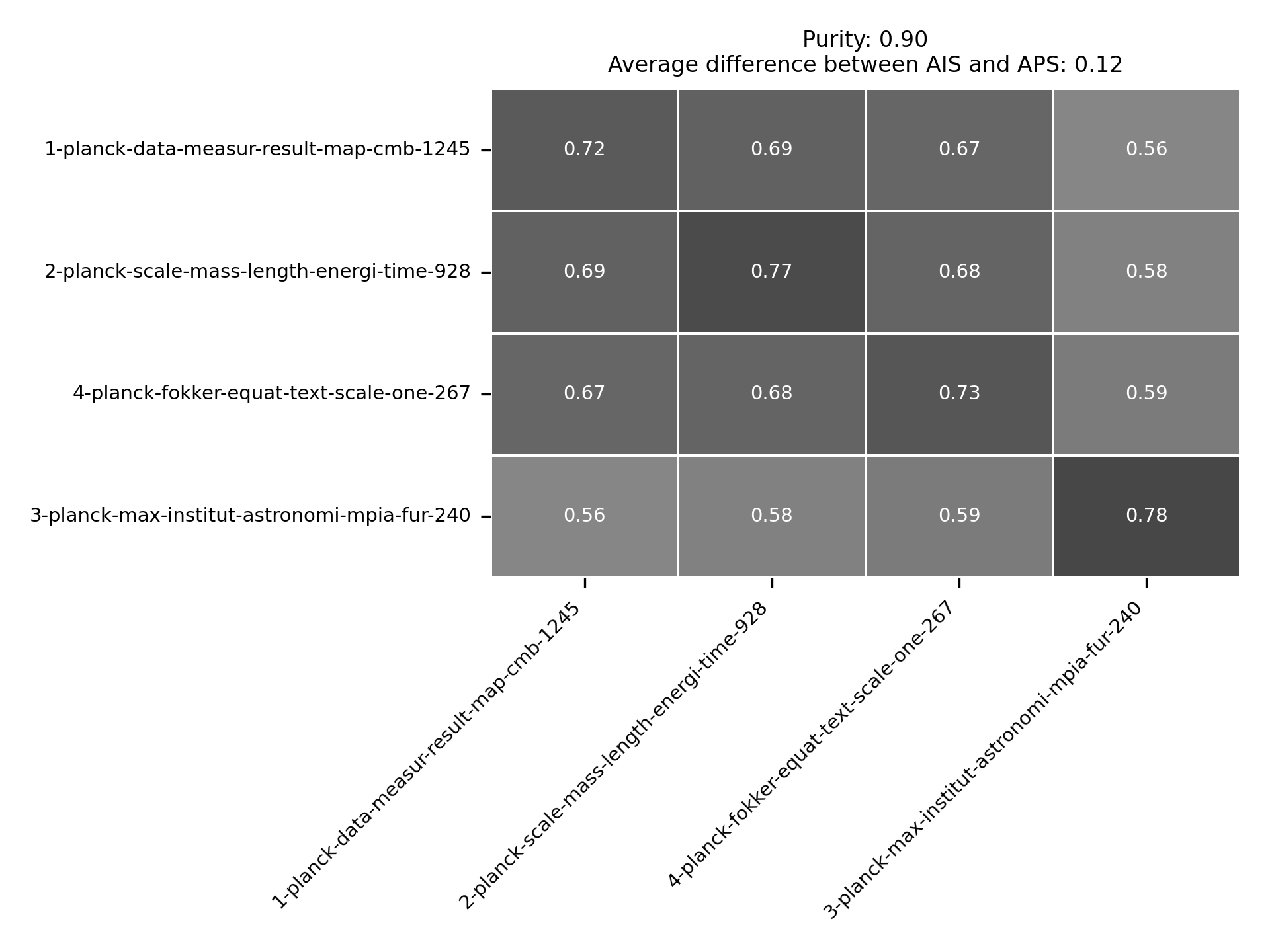}
        \caption{BERT}
    \end{subfigure}
    \caption{Heatmaps showing cluster cohesion and separation for each model's best four-label clustering solutions on the Astro-HEP-Planck Corpus. The diagonal cells display average inner similarity (AIS) scores for each cluster, reflecting internal cohesion, while the off-diagonal cells show average pairwise similarity (APS) between clusters, indicating separation. Clusters are sorted by size, with the largest appearing at the top and left. Cluster labels include the dominant label index, the six most frequent neighboring words within 10 tokens of ``Planck'', and the total number of embeddings in the cluster. Overall purity scores and the average differences between AIS and APS are noted above each heatmap.}
    \label{fig:cluster-examples}
\end{figure}

Figure \ref{fig:cluster-examples} presents similarity heatmaps for the best clustering solutions (based on inertia) of \texttt{Astro-HEP-BERT}, \texttt{astroBERT}, \texttt{PhysBERT}, and \texttt{BERT} in the four-label case of the Astro-HEP Corpus. This configuration is ideal for a focused analysis of separation and cohesion across models because it combines balanced complexity---with clear, interpretable heatmaps that highlight trends across models---and similarly high purity scores of 0.99 for \texttt{astroBERT}, \texttt{Astro-HEP-BERT}, and \texttt{PhysBERT}, while \texttt{SciBERT} and \texttt{BERT} perform noticeably lower. 

Each subfigure shows the mutual APS scores between the four clusters, with the AIS scores on the diagonal. For each clustering solution, the purity score as well as the average difference between AIS and APS scores are displayed at the top of the heatmap. Cluster names start with the index of the dominant label (as in Figure \ref{fig:cluster-comparison}), followed by the six most frequent words within 10 tokens on either side of the ``Planck'' occurrences (stop words excluded), and the number of embeddings in the cluster. The heatmaps are sorted by cluster size, with the largest clusters positioned at the top and left.

\texttt{Astro-HEP-BERT}, \texttt{PhysBERT}, and \texttt{astroBERT} all achieve peak and matching purity scores, followed by \texttt{SciBERT} and then \texttt{BERT}, which performs significantly weaker. A comparison of the heatmaps reveals notable variations in APS and AIS scores across models, reflecting how well they balance internal cluster cohesion and separation. To evaluate this balance, we can examine the difference between a model's AIS and APS scores. 

In the four-label case, \texttt{PhysBERT} and \texttt{Astro-HEP-BERT} achieve the best balance, with average differences of 0.33 and 0.32, respectively. \texttt{PhysBERT} demonstrates stronger coherence within its first two clusters, dominated by the \textit{MISSION} and \textit{UNITS} labels, though these clusters are less well-separated. In contrast, \texttt{Astro-HEP-BERT} has weaker coherence in its first two clusters but better separation, suggesting that they are more semantically diverse but also more distinct from each other. \texttt{astroBERT} achieves the highest AIS values overall (0.8 to 0.85), reflecting strong internal cohesion, but also has higher APS values, indicating less separation between clusters. For instance, clusters 1, 2, and 4 (recall that these indices reflect the actual frequency distribution of labels in the corpus rather than in the clustering solution) are much less distinct from one another in \texttt{astroBERT} compared to the other models. \texttt{SciBERT} shows very similar patterns, but performs worse than \texttt{astroBERT}. The weakest performer is clearly \texttt{BERT}, with smaller differences between AIS and APS, averaging 0.12. This indicates relatively poor separation between clusters, aligning with its lower purity scores and overall weaker performance in disambiguation.

As further detailed in the \textbf{Supplement} (\ref{sec:isotropy-analysis}), the improved balance between AIS and APS observed in \texttt{PhysBERT} and \texttt{Astro-HEP-BERT}, compared to \texttt{astroBERT} and the other two models, may only partially result from \textbf{differences in isotropy}---the degree to which embeddings are evenly distributed within the vector space. While \texttt{Astro-HEP-BERT} demonstrates the highest global isotropy among all models, \texttt{astroBERT} lies at the opposite end, exhibiting the most anisotropic embedding space. Nevertheless, \texttt{astroBERT} outperforms both \texttt{SciBERT} and \texttt{BERT} on our WSD and WSI tasks, despite their relatively higher isotropy.

Having established the models' ability to distinguish the different senses of ``Planck'' effectively, we now turn to their application in a diachronic setting, tracking how the distribution of these senses evolves over time. This analysis bridges sense disambiguation with the broader task of understanding conceptual shifts in scientific language.

\section{Lexical semantic change}
\label{sec:lexical-semantic-change}

Building on the models' tested performance in WSD and WSI from the previous sections, we can now explore their ability to measure \textbf{semantic change} for the term ``Planck'' \textbf{in a real-world, large-corpus setting}. By tracking the evolving distributions of CWEs across sense clusters, we examine changes over three decades in the unlabeled Astro-HEP Corpus, which includes over 300,000 instances of ``Planck''. A K-Means five-cluster solution, selected from 100 random seed initializations based on inertia, was applied to each model.

\begin{figure}[htbp!]
    \centering
    \begin{subfigure}[t]{.49\linewidth}
        \centering\includegraphics[width=1\linewidth]{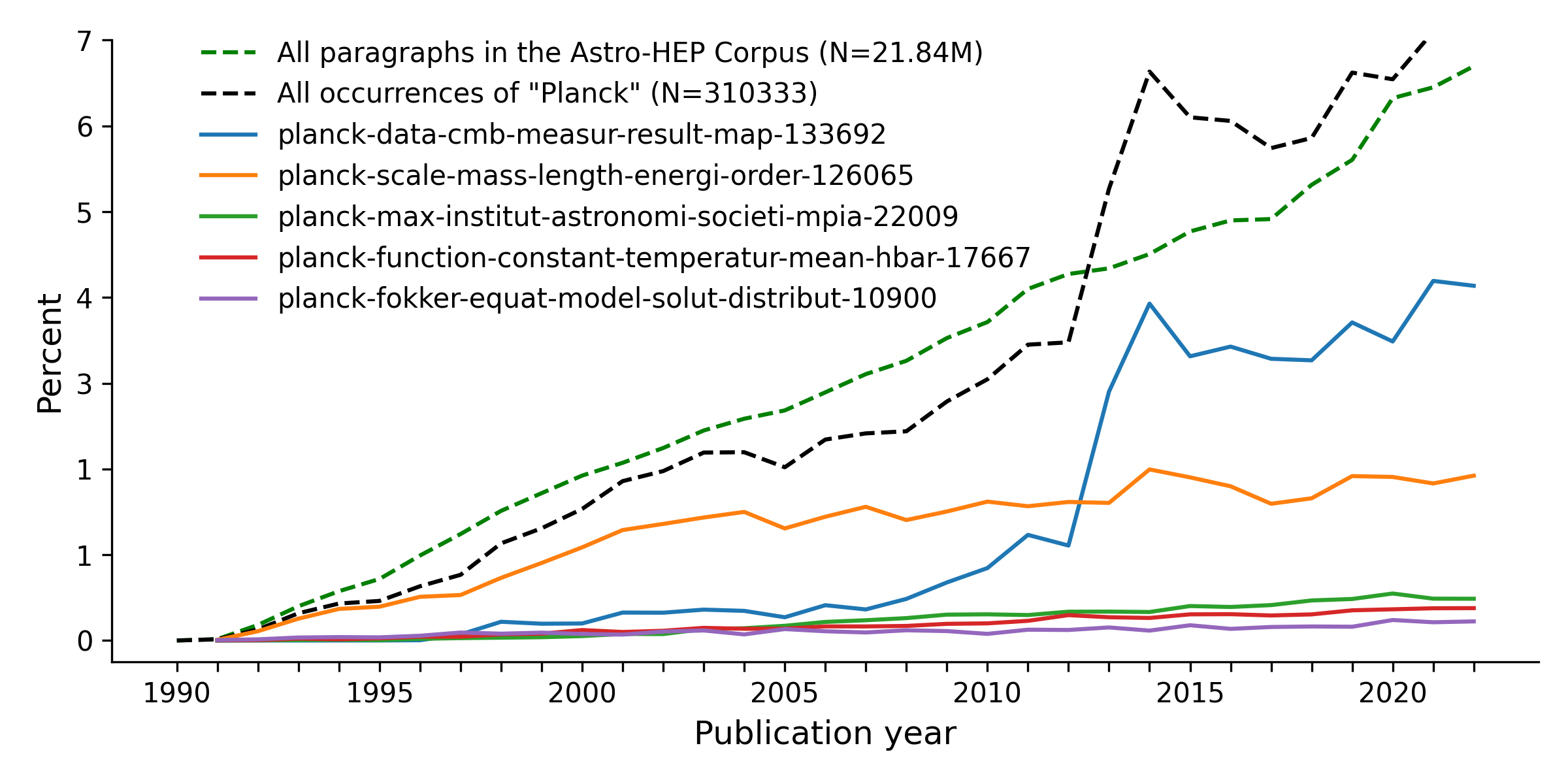}
        \caption{PhysBERT}
    \end{subfigure}
    \begin{subfigure}[t]{.49\linewidth}
        \centering\includegraphics[width=1\linewidth]{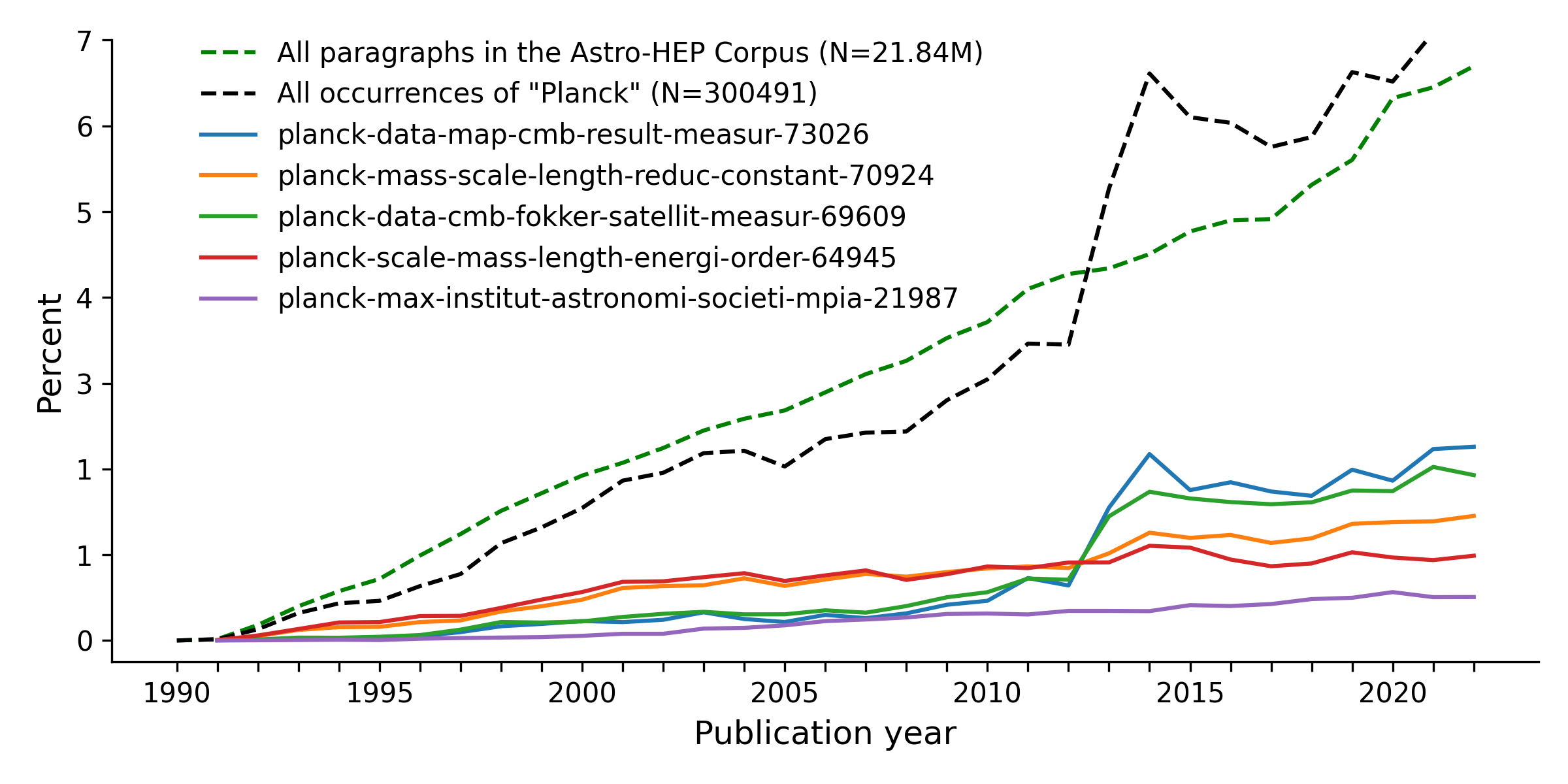}
        \caption{BERT}
    \end{subfigure}
    \caption{Evolution of the relative frequency of ``Planck'' occurrences across five clusters (colored lines) and over time, as modeled using \texttt{PhysBERT} (a) and \texttt{BERT} (b). The x-axis shows the years from 1990 to 2022, and the y-axis represents the normalized frequencies of occurrences per year. The dashed black line indicates the overall relative frequency of ``Planck'', while the dashed green line represents overall corpus growth. Each cluster label includes the six most frequent neighboring words and the total number of embeddings assigned to the cluster.}
    \label{fig:senses-per-year}
\end{figure}

Figures \ref{fig:senses-per-year}a and \ref{fig:senses-per-year}b illustrate the relative prominence of these clusters (colored lines) over time for \texttt{PhysBERT} and \texttt{BERT}, respectively. The figures also show the overall relative frequency of ``Planck'' occurrences (dashed black line) and corpus growth (dashed green line). As before, cluster names were derived from the six most frequent words found in a 10-token window around ``Planck'' (excluding stop words), accompanied by the number of embeddings per cluster. The focus on \texttt{PhysBERT} (a domain-specific model) and \texttt{BERT} (a general-purpose model) reflects their contrasting design and the relative quality of their clustering solutions.

\texttt{PhysBERT} (\ref{fig:senses-per-year}a) produced the most interpretable five-cluster breakdown, with clusters corresponding to prominent senses of ``Planck'': the Planck mission (blue), Planck units (orange), the Max Planck Society (green), Planck's law and constant (red), and the Fokker-Planck equation (purple). Notably, \texttt{PhysBERT} avoids splitting the mission and units senses into multiple clusters, yielding clearer and more distinct sense groupings. From 1996 to 1998, the (blue) mission cluster shows an initial rise, surpassing all clusters except the larger (orange) units cluster. Between 2007 and 2011, the mission cluster grows rapidly, overtaking the units cluster from 2012 to 2013. This growth peaks in 2014, after which the mission cluster stabilizes at a frequency roughly double that of the units cluster, maintaining this level through 2022.

In contrast, \texttt{BERT} (Figure \ref{fig:senses-per-year}b) disperses the mission and units senses across multiple clusters (blue/green for mission and orange/red for units), obscuring their relative growth. The model also blends the mission and Fokker-Planck equation senses in the green cluster, resulting in the least interpretable five-cluster breakdown. Still, the clusters referencing the mission sense (blue and green) surpass all others between 2012 and 2013.

These results align with milestones in the \textbf{history of the Planck mission}.\footnote{\url{https://www.esa.int/Science_Exploration/Space_Science/Planck\_overview};\\\url{https://www.cosmos.esa.int/web/planck/mission-history};\\\url{https://en.wikipedia.org/wiki/Planck\_(spacecraft)}} Originally proposed as the COBRAS/SAMBA mission in the early 1990s, it was approved by the European Space Agency (ESA) in 1996 and renamed the Planck mission in honor of Max Planck. Around this time, references to the mission began appearing in scientific literature. Later, from 2011 to 2018, ESA released key data from the mission, including the 2013 first all-sky map of the cosmic microwave background (CMB), which may have contributed to the prominence of the mission-related sense during this period.

To quantify changes in the sense distribution of ``Planck'' and potentially the term's overall meaning, I applied two complementary indicators described in \citet{periti_lexical_2024}. The first, \textbf{Jensen-Shannon divergence (JSD)}, is a ``sense-based'' metric that captures shifts in the relative prominence of distinct senses over time by comparing their normalized frequency distributions. It is particularly sensitive to changes in the balance of senses, even if the dominant meaning remains stable. JSD is calculated as:

$$\mbox{JSD}(d_1, d_2) = \frac{1}{2}(\mbox{KL}(d_1, M) + \mbox{KL}(d_2, M)),$$

where $d_1$ and $d_2$ are the frequency distributions of clusters in two consecutive years, $\mbox{KL}$ is the Kullback-Leibler divergence, and $M$ is the average of the two distributions.

In contrast, the second indicator, \textbf{cosine distance between prototypes (CDPT)}, is a ``form-based'' metric that tracks changes in the overall meaning of a term by comparing the averaged word embeddings (prototypes) from consecutive years. CDPT is less affected by fluctuations in sense distributions and instead highlights broader shifts in dominant meaning. It is defined as:

$$\mbox{CDPT}(\mbox{PT}_1, \mbox{PT}_2) = 1 - \mbox{CS}(\mbox{PT}_1, \mbox{PT}_2),$$

where $\mbox{CS}$ is the cosine similarity and $\mbox{PT}_1$ and $\mbox{PT}_2$ are the year-specific prototypes.

\begin{figure}[htbp!]
    \centering
    \begin{subfigure}[t]{.49\linewidth}
        \centering\includegraphics[width=1\linewidth]{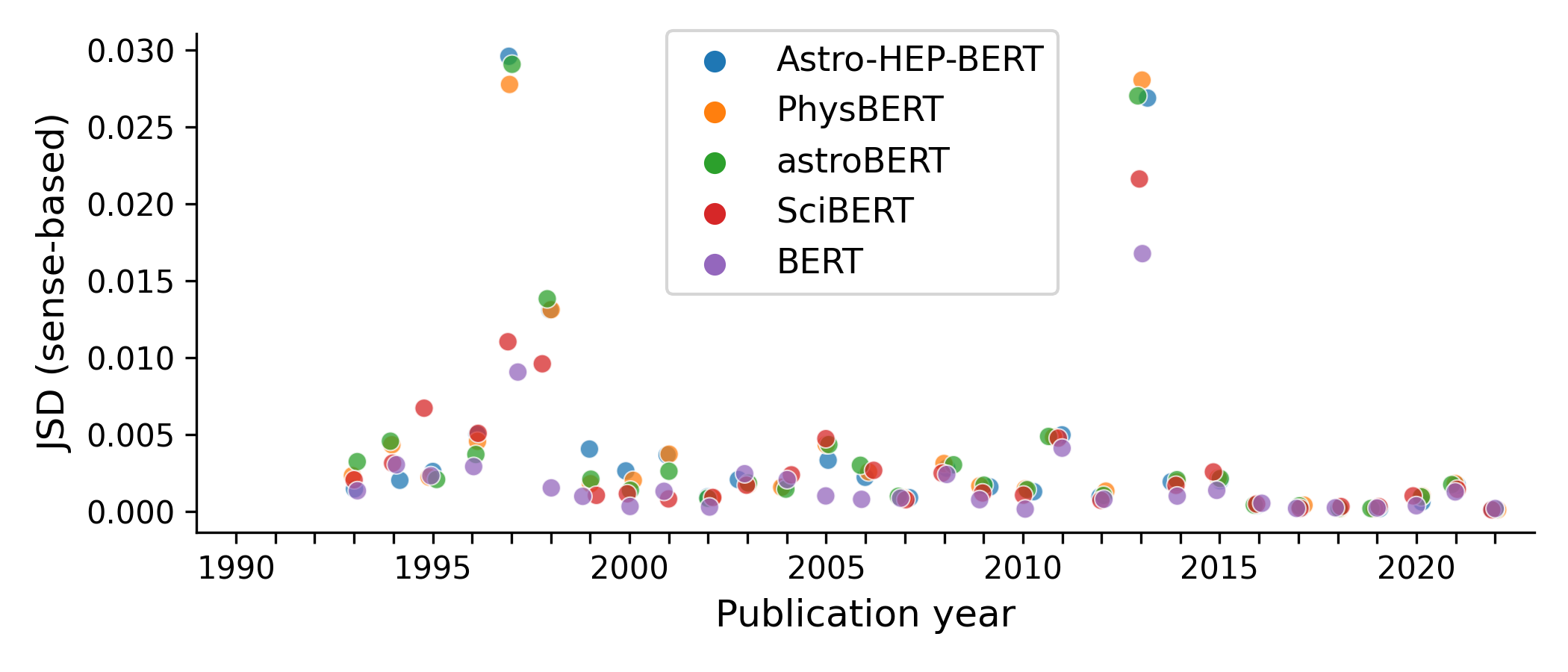}
        \caption{Jensen-Shannon divergence (JSD)}
    \end{subfigure}
    \begin{subfigure}[t]{.49\linewidth}
        \centering\includegraphics[width=1\linewidth]{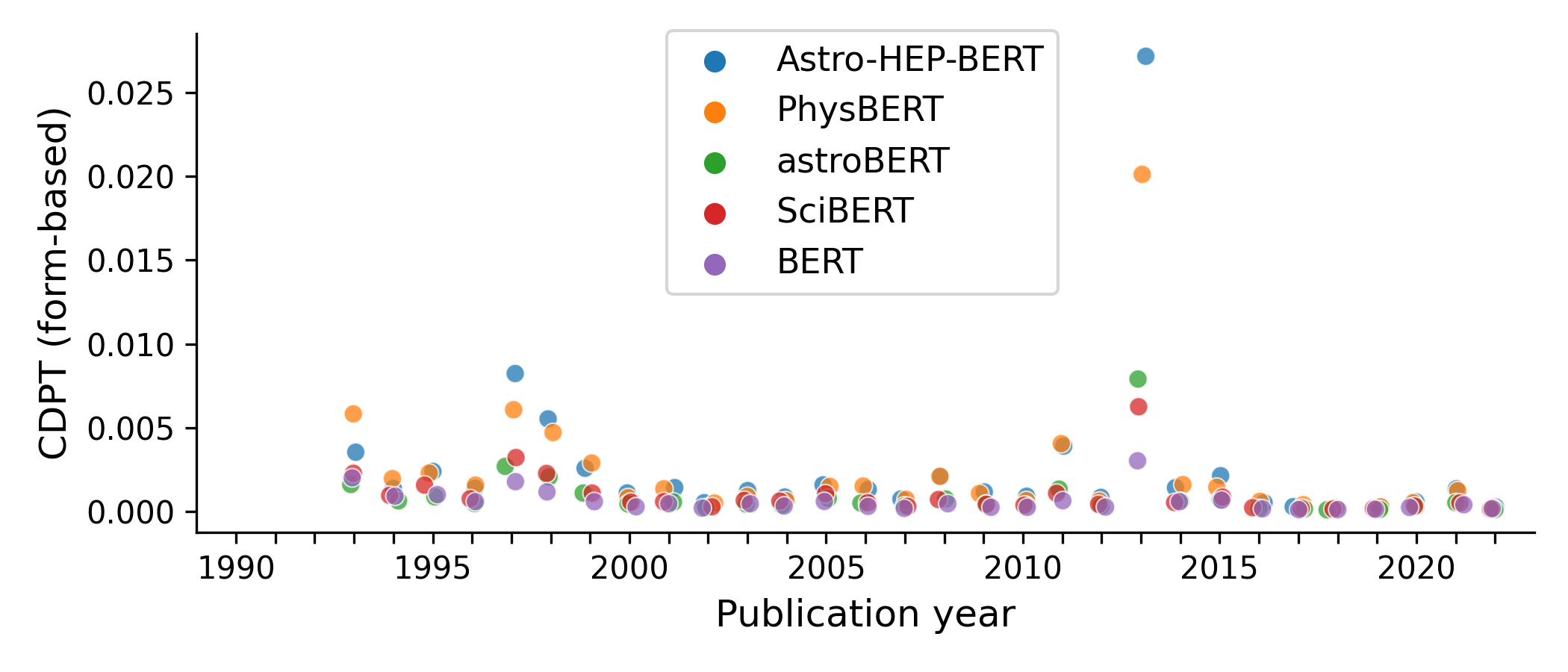}
        \caption{Cosine distance between prototypes (CDPT)}
    \end{subfigure}
    \caption{Semantic change of the term ``Planck'' over time, measured using two indicators: Jensen-Shannon divergence (JSD) and cosine distance between prototypes (CDPT). JSD (a) captures shifts in the distributions of clusters, representing different senses, while CDPT (b) measures global shifts in the aggregated meaning of the term. Both plots track changes across the years from 1990 to 2022 for each model.}
    \label{fig:JSD-and-CDPT}
\end{figure}

Figures \ref{fig:JSD-and-CDPT}a and \ref{fig:JSD-and-CDPT}b illustrate the results of JSD and CDPT, revealing significant semantic shifts in 1997 and 2013 that align with historical milestones of the Planck mission. In 1997, the renaming of the COBRAS/SAMBA mission to ``Planck'' coincides with a sharp spike in JSD, particularly in domain-specific models like \texttt{Astro-HEP-BERT} and \texttt{PhysBERT}, capturing the rise of the mission sense even as the dominant meaning remained tied to units. CDPT showed a smaller spike, suggesting limited impact on the overall dominant meaning. Similarly, in 2013, the release of key data from the Planck mission aligns with sharp increases in both indicators, reflecting a more cohesive shift in the dominant meaning of ``Planck''.

The complementary nature of JSD and CDPT is evident here: JSD is more sensitive to shifts in the balance of senses, such as the emergence of the mission sense in 1997, while CDPT highlights broader contextual changes, as seen in the major semantic shift in 2013. Post-2013, both indicators stabilize, but JSD captures ongoing smaller fluctuations in sense distributions, whereas CDPT reflects the settling of the dominant meaning.

These findings demonstrate \textbf{the value of combining sense-based and form-based LSC indicators} to capture both fine-grained sense dynamics and broader shifts in dominant meaning, providing a comprehensive view of lexical semantic change for scientific concepts over time. Among the models, domain-specific ones like \texttt{Astro-HEP-BERT} and \texttt{PhysBERT} display greater sensitivity to both JSD and CDPT changes, reflecting their superior performance in WSD and WSI tasks.

\section{Conclusion and discussion}
\label{sec:conclusion-and-discussion}

This study demonstrates the value of contextualized word embeddings (CWEs) for the history, philosophy, and sociology of science (HPSS), providing \textbf{a new tool for analyzing the meanings of scientific concepts}. Using the term ``Planck'' as a test case, I evaluated five BERT-based models across tasks in word sense disambiguation, sense induction, clustering quality, and lexical semantic change. The findings show that CWEs effectively disambiguate scientific terms, uncover distinct meanings, and trace diachronic trends, revealing how concepts like ``Planck'' evolve alongside developments in science, technology, and society.

The study highlights the promise of \textbf{domain-adapted CWEs for HPSS research}, particularly when models are trained on specialized corpora tailored to specific scientific fields or historical periods. This approach maximizes contextual relevance, and while critics may fear that such specialization risks overfitting, we can instead view it as a form of context-sensitive tailoring that aligns a model's interpretive focus with the unique conceptual landscape of its target domain. By allowing models to learn the distinct characteristics of specific scientific discourses, this approach enhances performance in these contexts, even at the expense of broader generalizability. In this regard, the success of \textbf{Astro-HEP-BERT serves as a proof-of-concept} for developing specialized models for HPSS on a limited budget. By reusing pretrained general-purpose weights, Astro-HEP-BERT achieved performance comparable to, or better than, more resource-intensive domain-adapted models like \texttt{PhysBERT} and \texttt{astroBERT}, despite using fewer training examples.

Beyond tracing ``conceptual histories'' \citep{zichert_tracing_2024, wevers_digital_2020}, domain-adapted CWEs and their basic operations---such as measuring similarity, clustering, and generating prototypes---may offer broader applications for exploring \textbf{other HPSS phenomena related to semantic similarity and divergence}. These could include investigating lexical codification within research fields \citep{teich_less_2021, glaser_scientific_2018} or addressing the complexities of translating scientific concepts across paradigms \citep{kuhn_last_2022} and within ``trading zones'' \citep{galison_image_1997}. For instance, semantic distance measures could reveal patterns of conceptual consistency, while clustering CWEs could uncover emergent terminologies or hybrid concepts bridging disciplines. Prototypical embeddings for terms from different paradigms could further illuminate semantic alignment or divergence, offering new and complementary methods for studying the evolution of scientific language. These possibilities remain speculative and sketchy, outlining potential approaches that must be explored further and specified through targeted studies and practical applications.

CWEs trained on scientific corpora can also \textbf{enhance downstream tasks and advanced computational workflows}, offering new opportunities for specialized applications in HPSS. For instance, they can improve domain-specific named entity recognition (NER) or the generation of specialized sentence embeddings, as demonstrated by recent studies \citep{hellert_physbert_2024, grezes_improving_2022}. Beyond these tasks, domain-adapted CWEs could play a pivotal role in advanced computational workflows. A compelling example is retrieval-augmented generation (RAG), where embedding domain-specific queries and documents into a shared semantic space enhances retrieval accuracy and contextual relevance. This integration could enable more precise information retrieval, facilitate robust knowledge synthesis, and support dynamic interactions with complex scientific corpora.

For future research, scholars should consider developing \textbf{HPSS-specific evaluation datasets} tailored to a variety of computational tasks. Foundational tasks like WSD, WSI, LSC, and Word-in-Context (WiC) \citep{pilehvar_wic_2019} remain essential. WiC, which tests whether a word maintains the same meaning across two different contexts, can also be extended to diachronic settings \citep{loureiro_tempowic_2022}, providing insights into the evolution of scientific concepts over time. Beyond these, the scope of evaluation datasets could be broadened to include tasks such as domain-specific NER and the generation of sentence embeddings using models like SentenceBERT \citep{reimers_sentence-bert_2019}. By creating tailored evaluation datasets for these and other tasks, researchers can address the unique challenges posed by specialized scientific languages, fostering more precise and context-aware computational analyses. My own two labeled datasets for the term ``Planck,'' which provide sense annotations based on predefined categories and, in one case---the Astro-HEP-Planck Corpus---include the publication year, exemplify how HPSS-specific annotations can support the classical WSD task and prepare for both synchronic and diachronic analyses.

A key limitation is that \textbf{BERT-based CWEs rely entirely on textual data}, which restricts analyses to patterns and contexts within the analyzed corpus. Non-textual aspects of scientific practice---such as experimental settings, interpersonal communication, and institutional or cultural contexts---are often inaccessible or only indirectly reflected in texts. To address this, HPSS researchers can combine CWEs with qualitative methods and explore multimodal pipelines. \textbf{Qualitative methods} provide the interpretive depth needed to analyze complex senses and socio-historical factors, ensuring CWE-based analyses remain grounded in broader historical and disciplinary contexts. \textbf{Multimodal pipelines}, by integrating non-textual sources like figures, diagrams, metadata, and citation networks, might help bridge the gap between textual representations and the material and social dimensions of scientific activity. However, the feasibility of such approaches remains to be fully explored \citep{yin_survey_2024}.

Staying informed about \textbf{advancements in adjacent fields}, such as computational humanities, computational linguistics, and machine learning, is essential for researchers employing CWEs in HPSS. By engaging with developments in model architectures, semantic tools, and advanced pipelines, HPSS scholars can enhance the precision and interpretability of CWE-based analyses. This interdisciplinary approach will not only deepen our understanding of scientific concepts and their evolution but also open new avenues for studying the dynamics of scientific language and its broader cultural implications. By embracing CWEs and their extensions, HPSS researchers have an opportunity to lead a transformative shift in the computational analysis of scientific discourse.

\section*{Acknowledgements}

I am grateful to my colleagues \textbf{Adrian Wüthrich} and \textbf{Michael Zichert} for their insightful feedback on this draft, and for our collaborative exchange of ideas regarding the use of computational methods in HPSS. I also want to acknowledge funding by \textbf{the European Union} (ERC Consolidator Grant, Project No. 101044932). Views and opinions expressed are however those of the author only and do not necessarily reflect those of the European Union or the European Research Council. Neither the European Union nor the granting authority can be held responsible for them.

\bibliographystyle{apalike} 
\bibliography{main}

\begin{thebibliography}{}

\bibitem[Ait-Saada and Nadif, 2023]{ait-saada_is_2023}
Ait-Saada, M. and Nadif, M. (2023).
\newblock Is anisotropy truly harmful? a case study on text clustering.
\newblock In Rogers, A., Boyd-Graber, J., and Okazaki, N., editors, {\em Proceedings of the 61st Annual Meeting of the Association for Computational Linguistics (Volume 2: Short Papers)}, pages 1194--1203. Association for Computational Linguistics.

\bibitem[Beltagy et~al., 2019]{beltagy_scibert_2019}
Beltagy, I., Lo, K., and Cohan, A. (2019).
\newblock {SciBERT}: A pretrained language model for scientific text.
\newblock {\em {arXiv}:1903.10676}.

\bibitem[Bevilacqua et~al., 2021]{bevilacqua_recent_2021}
Bevilacqua, M., Pasini, T., Raganato, A., and Navigli, R. (2021).
\newblock Recent trends in word sense disambiguation: A survey.
\newblock In {\em Proceedings of the Thirtieth International Joint Conference on Artificial Intelligence}, pages 4330--4338.

\bibitem[Biś et~al., 2021]{bis_too_2021}
Biś, D., Podkorytov, M., and Liu, X. (2021).
\newblock Too much in common: Shifting of embeddings in transformer language models and its implications.
\newblock In Toutanova, K., Rumshisky, A., Zettlemoyer, L., Hakkani-Tur, D., Beltagy, I., Bethard, S., Cotterell, R., Chakraborty, T., and Zhou, Y., editors, {\em Proceedings of the 2021 Conference of the North American Chapter of the Association for Computational Linguistics: Human Language Technologies}, pages 5117--5130. Association for Computational Linguistics.

\bibitem[Blei et~al., 2003]{blei_latent_2003}
Blei, D.~M., Ng, A.~Y., and Jordan, M.~I. (2003).
\newblock Latent dirichlet allocation.
\newblock {\em Journal of machine Learning research}, 3(Jan):993--1022.

\bibitem[Bowker and Star, 1999]{bowker_sorting_1999}
Bowker, G.~C. and Star, S.~L. (1999).
\newblock {\em Sorting things out: classification and its consequences}.
\newblock MIT Press.

\bibitem[Boyack et~al., 2011]{boyack_clustering_2011}
Boyack, K.~W., Newman, D., Duhon, R.~J., Klavans, R., Patek, M., Biberstine, J.~R., Schijvenaars, B., Skupin, A., Ma, N., and Börner, K. (2011).
\newblock Clustering more than two million biomedical publications: Comparing the accuracies of nine text-based similarity approaches.
\newblock {\em {PloS} one}, 6(3):e18029.

\bibitem[Cai et~al., 2020]{cai_isotropy_2020}
Cai, X., Huang, J., Bian, Y., and Church, K. (2020).
\newblock Isotropy in the contextual embedding space: Clusters and manifolds.
\newblock In {\em International conference on learning representations}.

\bibitem[Callon et~al., 1983]{callon_translations_1983}
Callon, M., Courtial, J.-P., Turner, W.~A., and Bauin, S. (1983).
\newblock From translations to problematic networks: An introduction to co-word analysis.
\newblock {\em Social Science Information}, 22(2):191 --235.

\bibitem[Canguilhem, 1991]{canguilhem_normal_1991}
Canguilhem, G. (1991).
\newblock {\em The Normal and the Pathological}.
\newblock MIT Press.

\bibitem[Chang, 2007]{chang_inventing_2007}
Chang, H. (2007).
\newblock {\em Inventing Temperature: Measurement and Scientific Progress}.
\newblock Oxford University Press.

\bibitem[Clarke et~al., 2010]{clarke_biomedicalization_2010}
Clarke, A.~E., Mamo, L., Fosket, J.~R., Fishman, J.~R., and Shim, J.~K., editors (2010).
\newblock {\em Biomedicalization: {Technoscience}, {Health}, and {Illness} in the {U}.{S}.}
\newblock Duke University Press.

\bibitem[Courtial and Law, 1989]{courtial_co-word_1989}
Courtial, J.-P. and Law, J. (1989).
\newblock A co-word study of artificial intelligence.
\newblock {\em Social Studies of Science}, 19(2):301 --311.

\bibitem[Daston and Galison, 2007]{daston_objectivity_2007}
Daston, L. and Galison, P. (2007).
\newblock {\em Objectivity}.
\newblock Zone Books.

\bibitem[Deerwester et~al., 1990]{deerwester_indexing_1990}
Deerwester, S., Dumals, S.~T., Furnas, G.~W., Landauer, K., T., and Harshman, R. (1990).
\newblock Indexing by latent semantic analysis.
\newblock {\em Journal of the American Society for Information Science}, 41(6):391--407.

\bibitem[Devlin et~al., 2018]{devlin_bert:_2018}
Devlin, J., Chang, M.-W., Lee, K., and Toutanova, K. (2018).
\newblock {BERT}: Pre-training of deep bidirectional transformers for language understanding.
\newblock {\em {arXiv}:1810.04805}.

\bibitem[Ding et~al., 2022]{ding_isotropy_2022}
Ding, Y., Martinkus, K., Pascual, D., Clematide, S., and Wattenhofer, R. (2022).
\newblock On isotropy calibration of transformer models.
\newblock In Tafreshi, S., Sedoc, J., Rogers, A., Drozd, A., Rumshisky, A., and Akula, A., editors, {\em Proceedings of the Third Workshop on Insights from Negative Results in {NLP}}, pages 1--9. Association for Computational Linguistics.

\bibitem[Ethayarajh, 2019]{ethayarajh_how_2019}
Ethayarajh, K. (2019).
\newblock How contextual are contextualized word representations? comparing the geometry of {BERT}, {ELMo}, and {GPT}-2 embeddings.
\newblock {\em {arXiv}:1909.00512}.

\bibitem[Fleck, 1979]{fleck_genesis_1979}
Fleck, L. (1979).
\newblock {\em Genesis and development of a scientific fact}.
\newblock University of Chicago Press.

\bibitem[Foucault, 1970]{foucault_order_1994}
Foucault, M. (1970).
\newblock {\em The Order of Things: An Archaeology of Human Sciences}.
\newblock Random House.

\bibitem[Galison, 1997]{galison_image_1997}
Galison, P. (1997).
\newblock {\em Image and Logic: A Material Culture of Microphysics}.
\newblock Univ of Chicago Pr, illustrated edition edition.

\bibitem[Gläser et~al., 2018]{glaser_scientific_2018}
Gläser, J., Laudel, G., Grieser, C., and Meyer, U. (2018).
\newblock Scientific fields as epistemic regimes: new opportunities for comparative science studies.
\newblock {\em TUTS Working Papers, 3-2018}.

\bibitem[Godey et~al., 2023]{godey_anisotropy_2023}
Godey, N., de~la Clergerie, E., and Sagot, B. (2023).
\newblock Is anisotropy inherent to transformers?
\newblock {\em {arXiv}:2306.07656}.

\bibitem[Grezes et~al., 2022]{grezes_improving_2022}
Grezes, F., Allen, T., Blanco-Cuaresma, S., Accomazzi, A., Kurtz, M.~J., Shapurian, G., Henneken, E., Grant, C.~S., Thompson, D.~M., Hostetler, T.~W., Templeton, M.~R., Lockhart, K.~E., Chen, S., Koch, J., Jacovich, T., and Protopapas, P. (2022).
\newblock Improving {astroBERT} using semantic textual similarity.
\newblock {\em {arXiv}:2212.00744}.

\bibitem[Grezes et~al., 2021]{grezes_building_2021}
Grezes, F., Blanco-Cuaresma, S., Accomazzi, A., Kurtz, M.~J., Shapurian, G., Henneken, E., Grant, C.~S., Thompson, D.~M., Chyla, R., {McDonald}, S., Hostetler, T.~W., Templeton, M.~R., Lockhart, K.~E., Martinovic, N., Chen, S., Tanner, C., and Protopapas, P. (2021).
\newblock Building {astroBERT}, a language model for astronomy \& astrophysics.
\newblock {\em {arXiv}:2112.00590}.

\bibitem[Gu et~al., 2021]{gu_domain-specific_2021}
Gu, Y., Tinn, R., Cheng, H., Lucas, M., Usuyama, N., Liu, X., Naumann, T., Gao, J., and Poon, H. (2021).
\newblock Domain-specific language model pretraining for biomedical natural language processing.
\newblock {\em {ACM} Transactions on Computing for Healthcare ({HEALTH})}, 3(1):1--23.

\bibitem[Hacking, 1975]{hacking_emergence_1975}
Hacking, I. (1975).
\newblock {\em The emergence of probability: A philosophical study of early ideas about probability, induction and statistical inference}.
\newblock Cambridge University Press.

\bibitem[Hacking, 1999]{hacking_social_1999}
Hacking, I. (1999).
\newblock {\em The Social Construction of What?}
\newblock Harvard University Press.

\bibitem[Hellert et~al., 2024]{hellert_physbert_2024}
Hellert, T., Montenegro, J., and Pollastro, A. (2024).
\newblock {PhysBERT}: {A} {Text} {Embedding} {Model} for {Physics} {Scientific} {Literature}.
\newblock {\em {arXiv}:2408.09574}.

\bibitem[Kleymann et~al., 2022]{kleymann_conceptual_2022}
Kleymann, R., Niekler, A., and Burghardt, M. (2022).
\newblock Conceptual forays: A corpus-based study of “theory” in digital humanities journals.
\newblock {\em Journal of Cultural Analytics}, 7(4).

\bibitem[Kuhn, 1962]{kuhn_structure_1962}
Kuhn, T.~S. (1962).
\newblock {\em The structure of scientific revolutions}.
\newblock University of Chicago Press.

\bibitem[Kuhn, 2022]{kuhn_last_2022}
Kuhn, T.~S. (2022).
\newblock {\em The Last Writings of Thomas S. Kuhn: Incommensurability in Science}.
\newblock University of Chicago Press.

\bibitem[Latour, 1987]{latour_science_1987}
Latour, B. (1987).
\newblock {\em Science in action: How to follow scientists and engineers through society}.
\newblock Harvard University Press.

\bibitem[Laubichler et~al., 2019]{laubichler_computational_2019}
Laubichler, M.~D., Maienschein, J., and Renn, J. (2019).
\newblock Computational history of knowledge: Challenges and opportunities.
\newblock {\em Isis}, 110(3):502--512.

\bibitem[Lean et~al., 2023]{lean_digital_2023}
Lean, O.~M., Rivelli, L., and Pence, C.~H. (2023).
\newblock Digital literature analysis for empirical philosophy of science.
\newblock {\em The British Journal for the Philosophy of Science}, 74(4):875--898.

\bibitem[Lee et~al., 2019]{lee_biobert_2019}
Lee, J., Yoon, W., Kim, S., Kim, D., Kim, S., So, C.~H., and Kang, J. (2019).
\newblock {BioBERT}: a pre-trained biomedical language representation model for biomedical text mining.
\newblock {\em {arXiv}:1901.08746}.

\bibitem[Leydesdorff and Rafols, 2009]{leydesdorff_global_2009}
Leydesdorff, L. and Rafols, I. (2009).
\newblock A global map of science based on the {ISI} subject categories.
\newblock {\em Journal of the American Society for Information Science and Technology}, 60(2):348--362.

\bibitem[Liu et~al., 2019]{liu_roberta_2019}
Liu, Y., Ott, M., Goyal, N., Du, J., Joshi, M., Chen, D., Levy, O., Lewis, M., Zettlemoyer, L., and Stoyanov, V. (2019).
\newblock {RoBERTa}: A robustly optimized {BERT} pretraining approach.
\newblock {\em {arXiv}:1907.11692}.

\bibitem[Loureiro et~al., 2022]{loureiro_tempowic_2022}
Loureiro, D., D'Souza, A., Muhajab, A.~N., White, I.~A., Wong, G., Anke, L.~E., Neves, L., Barbieri, F., and Camacho-Collados, J. (2022).
\newblock {TempoWiC}: An evaluation benchmark for detecting meaning shift in social media.
\newblock {\em {arXiv}:2209.07216}.

\bibitem[Loureiro et~al., 2020]{loureiro_language_2020}
Loureiro, D., Rezaee, K., Pilehvar, M.~T., and Camacho-Collados, J. (2020).
\newblock Language models and word sense disambiguation: An overview and analysis.
\newblock {\em {arXiv}:2008.11608}.

\bibitem[Malaterre and Léonard, 2024]{malaterre_epistemic_2024}
Malaterre, C. and Léonard, M. (2024).
\newblock Epistemic markers in the scientific discourse.
\newblock {\em Philosophy of Science}, 91(1):151--174.

\bibitem[Merchant, 1980]{merchant_death_1980}
Merchant, C. (1980).
\newblock {\em The Death of Nature: Women, Ecology, and the Scientific Revolution}.
\newblock Harper \& Row.

\bibitem[Mickus et~al., 2024]{mickus_isotropy_2024}
Mickus, T., Grönroos, S.-A., and Attieh, J. (2024).
\newblock Isotropy, clusters, and classifiers.
\newblock {\em {arXiv}:2402.03191}.

\bibitem[Mikolov et~al., 2013]{mikolov_efficient_2013}
Mikolov, T., Chen, K., Corrado, G., and Dean, J. (2013).
\newblock Efficient estimation of word representations in vector space.
\newblock {\em {arXiv}:1301.3781}.

\bibitem[Mol, 2002]{mol_body_2002}
Mol, A. (2002).
\newblock {\em The Body Multiple: Ontology in Medical Practice}.
\newblock Duke University Press.

\bibitem[Mu et~al., 2018]{mu_all_2018}
Mu, J., Bhat, S., and Viswanath, P. (2018).
\newblock All-but-the-top: Simple and effective postprocessing for word representations.
\newblock {\em {arXiv}:1702.01417}.

\bibitem[Overton, 2013]{overton_explain_2013}
Overton, J.~A. (2013).
\newblock “explain” in scientific discourse.
\newblock {\em Synthese}, 190(8):1383--1405.

\bibitem[Pence and Ramsey, 2018]{pence_how_2018}
Pence, C.~H. and Ramsey, G. (2018).
\newblock How to do digital philosophy of science.
\newblock {\em Philosophy of Science}, 85(5):930--941.

\bibitem[Periti and Montanelli, 2024]{periti_lexical_2024}
Periti, F. and Montanelli, S. (2024).
\newblock Lexical semantic change through large language models: a survey.
\newblock {\em ACM Computing Surveys}, 56(11):282:1--282:38.

\bibitem[Pickering, 1995]{pickering_mangle_1995}
Pickering, A. (1995).
\newblock {\em The Mangle of Practice: Time, Agency, and Science}.
\newblock University of Chicago Press.

\bibitem[Pilehvar and Camacho-Collados, 2019]{pilehvar_wic_2019}
Pilehvar, M.~T. and Camacho-Collados, J. (2019).
\newblock {WiC}: the word-in-context dataset for evaluating context-sensitive meaning representations.
\newblock In Burstein, J., Doran, C., and Solorio, T., editors, {\em Proceedings of the 2019 Conference of the North American Chapter of the Association for Computational Linguistics: Human Language Technologies, Volume 1 (Long and Short Papers)}, pages 1267--1273. Association for Computational Linguistics.

\bibitem[Reimers and Gurevych, 2019]{reimers_sentence-bert_2019}
Reimers, N. and Gurevych, I. (2019).
\newblock Sentence-{BERT}: Sentence embeddings using siamese {BERT}-networks.
\newblock {\em {arXiv}:1908.10084}.

\bibitem[Rip and Courtial, 1984]{rip_co-word_1984}
Rip, A. and Courtial, J.~P. (1984).
\newblock Co-word maps of biotechnology: An example of cognitive scientometrics.
\newblock {\em Scientometrics}, 6(6):381--400.

\bibitem[Simons, 2024]{simons_astro_2024}
Simons, A. (2024).
\newblock Astro-hep-bert: A bidirectional language model for studying the meanings of concepts in astrophysics and high energy physics.
\newblock {\em {arXiv}:}.

\bibitem[Steinle, 2016]{steinle_exploratory_2016}
Steinle, F. (2016).
\newblock {\em Exploratory Experiments: Ampère, Faraday, and the Origins of Electrodynamics}.
\newblock University of Pittsburgh Press.

\bibitem[Sun and Platos, 2023]{sun_method_2023}
Sun, Y. and Platos, J. (2023).
\newblock A method for constructing word sense embeddings based on word sense induction.
\newblock {\em Scientific Reports}, 13(1):12945.

\bibitem[Tahmasebi and Dubossarsky, 2023]{tahmasebi_computational_2023}
Tahmasebi, N. and Dubossarsky, H. (2023).
\newblock Computational modeling of semantic change.
\newblock {\em {arXiv}:2304.06337}.

\bibitem[Teich et~al., 2021]{teich_less_2021}
Teich, E., Fankhauser, P., Degaetano-Ortlieb, S., and Bizzoni, Y. (2021).
\newblock Less is more/more diverse: On the communicative utility of linguistic conventionalization.
\newblock {\em Frontiers in Communication}, 5.

\bibitem[Venturini et~al., 2014]{venturini_three_2014}
Venturini, T., Baya~Laffite, N., Cointet, J.-P., Gray, I., Zabban, V., and De~Pryck, K. (2014).
\newblock Three maps and three misunderstandings: A digital mapping of climate diplomacy.
\newblock {\em Big Data \& Society}, 1(2).

\bibitem[Wang et~al., 2020]{wang_improving_2020}
Wang, L., Huang, J., Huang, K., Hu, Z., Wang, G., and Gu, Q. (2020).
\newblock Improving neural language generation with spectrum control.
\newblock In {\em Proceedings of the International Conference on Learning Representations 2020}.

\bibitem[Wevers and Koolen, 2020]{wevers_digital_2020}
Wevers, M. and Koolen, M. (2020).
\newblock Digital begriffsgeschichte: Tracing semantic change using word embeddings.
\newblock {\em Historical Methods: A Journal of Quantitative and Interdisciplinary History}, 53(4):226--243.

\bibitem[Wiedemann et~al., 2019]{wiedemann_does_2019}
Wiedemann, G., Remus, S., Chawla, A., and Biemann, C. (2019).
\newblock Does {BERT} make any sense? interpretable word sense disambiguation with contextualized embeddings.
\newblock {\em {arXiv}:1909.10430}.

\bibitem[Yin et~al., 2024]{yin_survey_2024}
Yin, S., Fu, C., Zhao, S., Li, K., Sun, X., Xu, T., and Chen, E. (2024).
\newblock A survey on multimodal large language models.
\newblock {\em {arXiv}:2306.13549}.

\bibitem[Zichert and Wüthrich, 2024]{zichert_tracing_2024}
Zichert, M. and Wüthrich, A. (2024).
\newblock Tracing the development of the virtual particle concept using semantic change detection.
\newblock {\em {arXiv}:2410.16855}.

\end{thebibliography}

\section*{Supplement}
\setcounter{subsection}{0}
\setcounter{figure}{0}
\makeatletter 
\renewcommand{\thefigure}{S\@arabic\c@figure}
\renewcommand{\thesubsection}{S\@arabic\c@subsection}
\makeatother

\subsection{Isotropy analysis}
\label{sec:isotropy-analysis}

In Section \ref{sec:separation-and-cohesion} I hypothesized that the improved balance between AIS and APS observed in \texttt{PhysBERT} and \texttt{Astro-HEP-BERT}, compared to \texttt{astroBERT} and the other two models may be partly due to a more isotropic embedding space, where embeddings are more evenly distributed, allowing for better cluster separation. To investigate this further, this section begins with a brief overview of the ongoing debate in the literature regarding the roles of anisotropy and isotropy in BERT models and their impact on model performance.

The discussion around anisotropy in BERT models originated with the surprising discovery by \citet{ethayarajh_how_2019} that CWEs produced by BERT exhibit significant anisotropy---an uneven distribution of CWEs in the high-dimensional space. This finding contrasted with earlier research by \citet{mu_all_2018}, which demonstrated that anisotropy in \textit{static} word embeddings could lead to performance degradation. However, since BERT's \textit{contextualized} representations have shown remarkable performance across numerous tasks, researchers began questioning whether anisotropy had the same negative effect on contextualized embeddings as it did on static ones.

Subsequent studies produced mixed results regarding the impact of reducing anisotropy in BERT and other Transformer models. Some researchers reported performance improvements by mitigating anisotropy, using methods such as post-processing techniques \citep{bis_too_2021, wang_improving_2020} or modifications to pretraining objectives \citep{godey_anisotropy_2023}. However, others found no significant improvements or, in some cases, even performance declines when attempting to enforce isotropy \citep{ait-saada_is_2023, ding_isotropy_2022}. This has led to varying opinions on whether reducing anisotropy consistently enhances the performance of Transformer models across different tasks.

A related discussion has emerged around the distinction between global anisotropy and local isotropy. Several authors found that while Transformer models, including BERT, display global anisotropy (an uneven embedding distribution across the entire space), they show local isotropy within specific subspaces or clusters \citep{mickus_isotropy_2024, ait-saada_is_2023, ding_isotropy_2022, cai_isotropy_2020}. This local isotropy may explain BERT's high performance, as it allows for expressive representations in particular regions of the embedding space, despite the global anisotropy. Thus, the relationship between (an)isotropy and model performance, particularly for contextualized embeddings, remains complex and multifaceted.

The aim of this supplementary section is not to make a direct contribution to this broader debate, but rather to assess how our five models compare in terms of their global isotropy, which I calculated using the average cosine similarity (ACS) between the contextualized word embeddings (CWEs) of randomly sampled words---a standard approach in this research area. For random vectors with independent, identically distributed components, expected cosine similarity approaches zero as dimensionality increases. Thus, ACS values closer to zero suggest a more isotropic (and thus more balanced) embedding space.

For each model and corpus, 200,000 tokens were sampled, paired, and cosine similarity was calculated for 100,000 random pairs, with the resulting ACS providing a measure of global isotropy.

\begin{figure}
  \centering
  \begin{subfigure}[t]{.49\linewidth}
    \centering\includegraphics[width=1\linewidth]{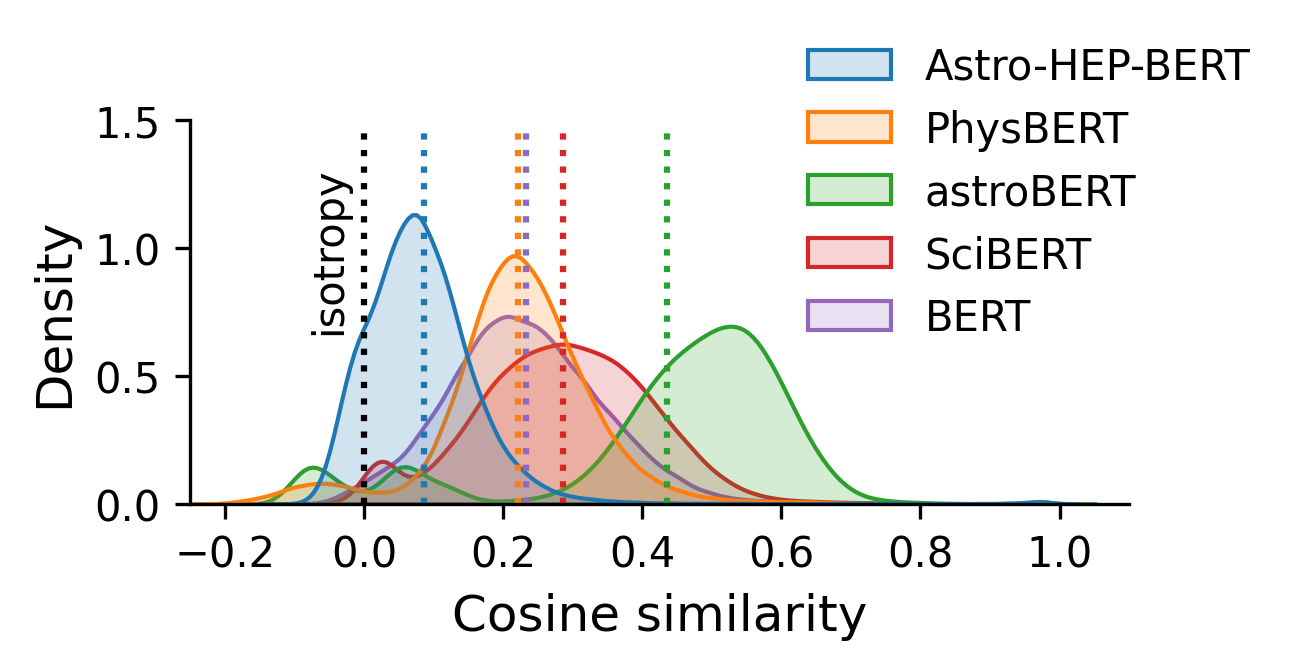}
    \caption{Astro-HEP Corpus (all models)}
  \end{subfigure}
  \begin{subfigure}[t]{.49\linewidth}
    \centering\includegraphics[width=1\linewidth]{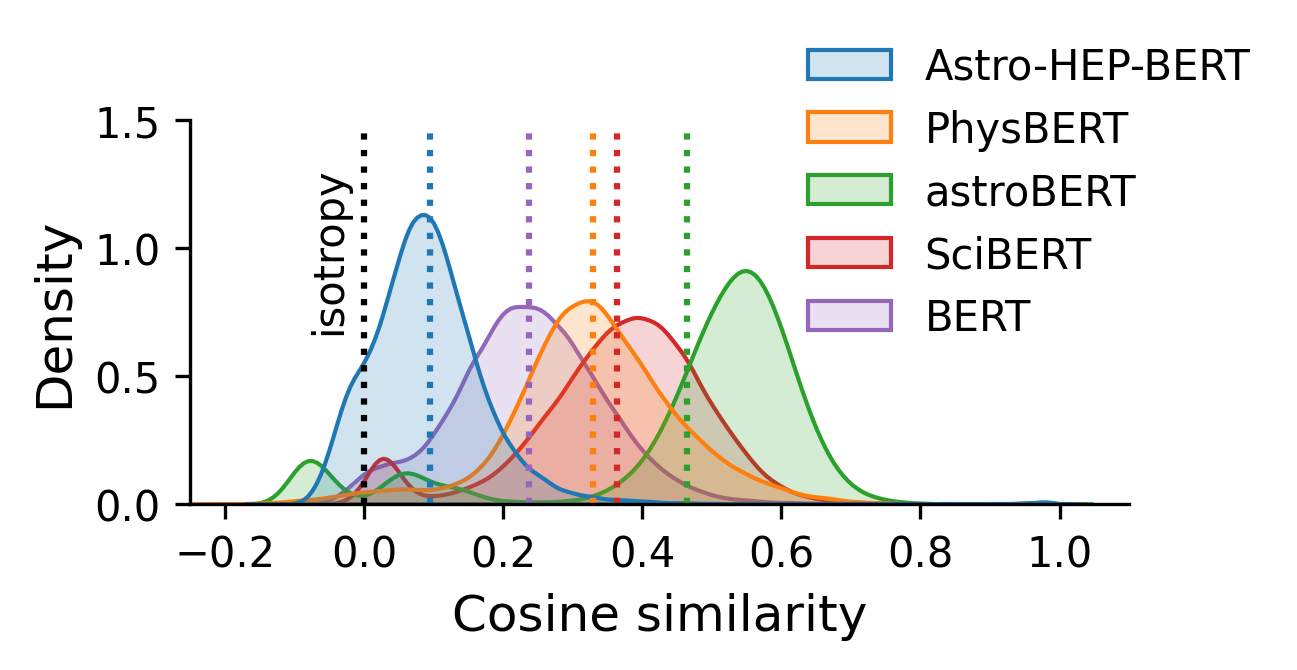}
    \caption{Wikipedia-Physics Corpus (all models)}
  \end{subfigure}
  \begin{subfigure}[t]{.49\linewidth}
    \centering\includegraphics[width=1\linewidth]{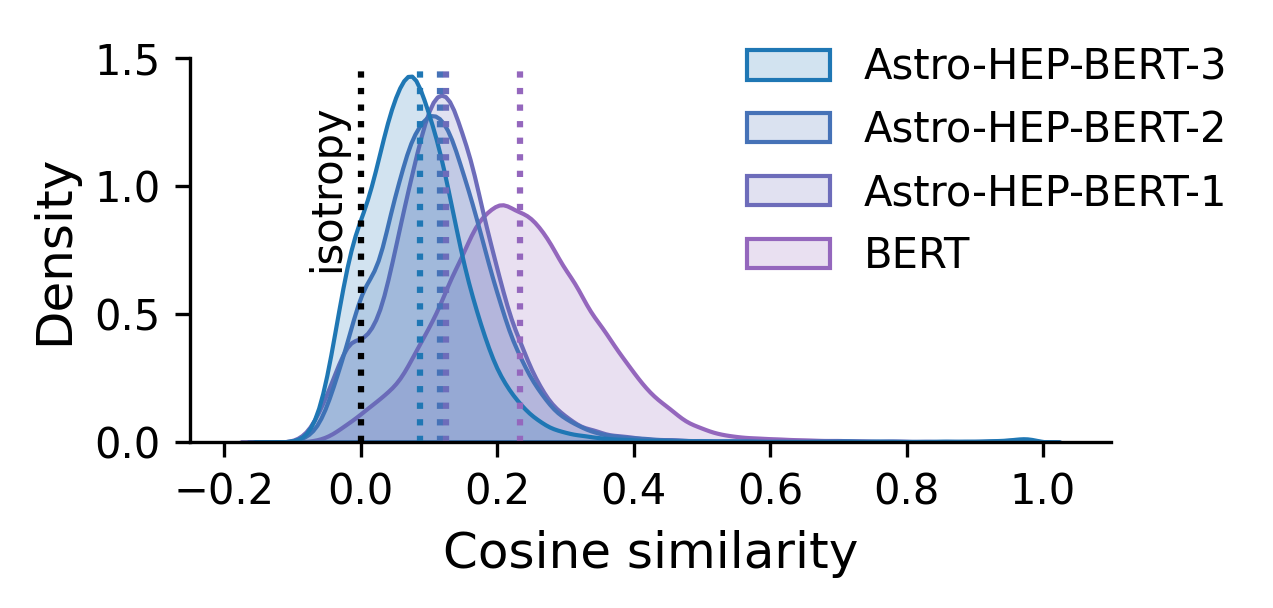}
    \caption{Astro-HEP Corpus (evolution of \texttt{Astro-HEP-BERT})}
  \end{subfigure}
  \begin{subfigure}[t]{.49\linewidth}
    \centering\includegraphics[width=1\linewidth]{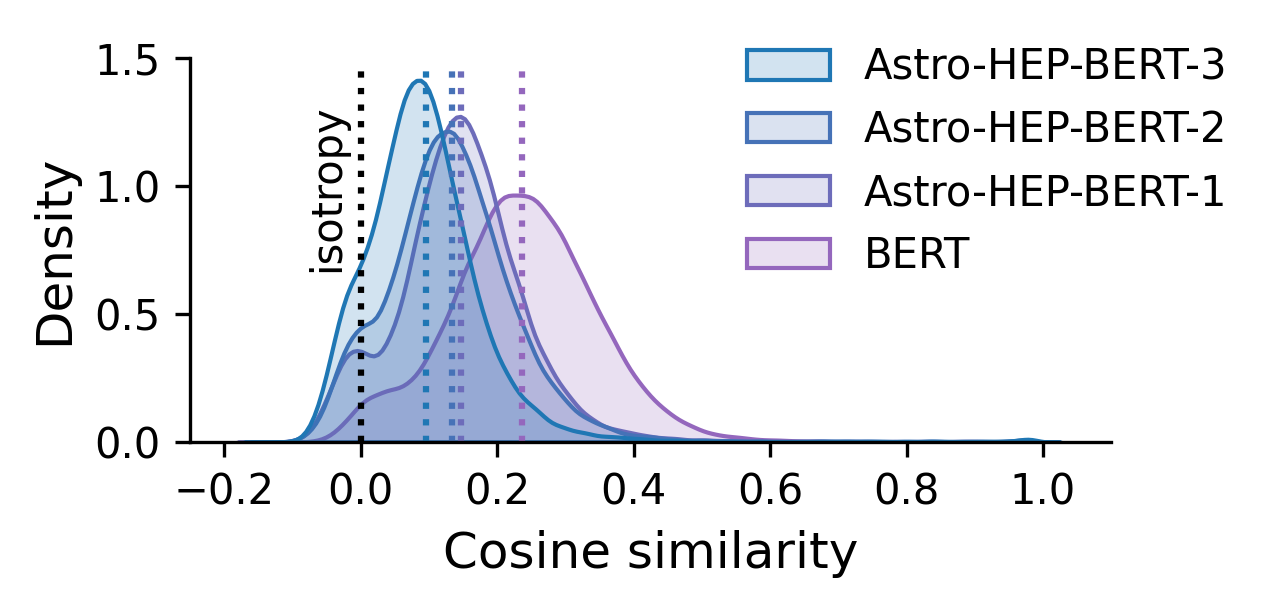}
    \caption{Wikipedia-Physics Corpus (evolution of \texttt{Astro-HEP-BERT})}
  \end{subfigure}
  \caption{Global isotropy analysis across models and corpora using average cosine similarity (ACS) and density distributions. (a) ACS and cosine similarity distribution for CWEs in the Astro-HEP Corpus, with each model's ACS represented by a colored vertical line. (b) ACS and cosine similarity distribution in the Wikipedia-Physics Corpus. (c, d) Evolution of ACS for \texttt{Astro-HEP-BERT} over three pretraining epochs, evaluated on the Astro-HEP (c) and Wikipedia-Physics (d) corpora, showing progressive improvement in isotropy. Lower ACS values indicate a more isotropic and balanced embedding space.}
  \label{fig:global-isotropy}
\end{figure}

Results indicate that \texttt{Astro-HEP-BERT} has the most isotropic embedding space, with an ACS just below 0.1 in both corpora, followed by \texttt{BERT} and \texttt{PhysBERT}, which show ACS values slightly above 0.2 in the Astro-HEP Corpus. In contrast, \texttt{PhysBERT} shows a slightly higher ACS in the Wikipedia-Physics Corpus, nearing 0.3. \texttt{SciBERT} and \texttt{astroBERT} demonstrate higher anisotropy, with \texttt{astroBERT} reaching an ACS near 0.5 in the Wikipedia-Physics Corpus.

Certain models, like \texttt{PhysBERT} and \texttt{SciBERT}, display multiple peaks in their cosine similarity distributions, while \texttt{Astro-HEP-BERT} and \texttt{BERT} exhibit a more singular peak. This difference suggests that models with multiple peaks might have more compartmentalized embedding spaces, indicating anisotropic regions or clustering tendencies, whereas \texttt{Astro-HEP-BERT}'s single, concentrated peak reflects a more isotropic, evenly distributed space.

Figures \ref{fig:global-isotropy}c and \ref{fig:global-isotropy}d show how isotropy evolved in \texttt{Astro-HEP-BERT} over three pretraining epochs. Notably, after the first epoch, ACS decreased by approximately 0.1, with smaller reductions in subsequent epochs, indicating progressive isotropy improvement as embeddings spread more uniformly.

These findings highlight two key observations: \texttt{Astro-HEP-BERT}'s increased isotropy relative to models like \texttt{PhysBERT} and \texttt{astroBERT} may stem from its adaptation of BERT's pretrained embedding space rather than starting from scratch. Refining BERT's embeddings with astrophysics and HEP-specific data likely preserved and enhanced isotropy by organizing domain-specific terms within an existing, well-structured framework. In contrast, models trained from scratch on physics texts appear to have developed more clustered or anisotropic embedding spaces, potentially due to early overfitting on domain-specific patterns without a generalized framework.

\end{document}